\documentclass[11pt]{article}
\usepackage[margin=1in]{geometry}
\usepackage[T1]{fontenc}
\usepackage{times}
\usepackage[round]{natbib}
\usepackage{authblk}
\usepackage{graphicx}
\usepackage{xcolor}
 
\usepackage{amsmath}                 
\usepackage[normalem]{ulem}          
 
\newcommand{\added}[1]{#1}
\newcommand{\deleted}[1]{}

\usepackage{amsmath,amsfonts,bm}

\def\eqref#1{equation~\ref{#1}}

\def\1{\bm{1}}

\DeclareMathAlphabet{\mathsfit}{\encodingdefault}{\sfdefault}{m}{sl}
\SetMathAlphabet{\mathsfit}{bold}{\encodingdefault}{\sfdefault}{bx}{n}

\usepackage{hyperref}
\usepackage{url}
\usepackage{booktabs}
\usepackage{array}
\usepackage{makecell}
\usepackage{listings}
\usepackage{tabularx}
 
\definecolor{cfgkey}{RGB}{25,50,130}
\definecolor{cfgval}{RGB}{150,70,20}
 
\lstdefinelanguage{cfglang}{
  morekeywords={true,false,null,empty},
  sensitive=false,
  morecomment=[l]{:},
  morestring=[b]",
}
\usepackage{textcomp}   
\usepackage{upquote}
\lstdefinestyle{cfg}{
  language=cfglang,
  basicstyle=\ttfamily\scriptsize,
  keywordstyle=\color{cfgval}\bfseries,
  stringstyle=\color{cfgkey},
  commentstyle=\color{cfgval},
  showstringspaces=false,
  columns=fullflexible,
  keepspaces=true,
  aboveskip=0.6\baselineskip,
  belowskip=0.6\baselineskip,
}
 
\title{PRETZEL: A Benchmark for Cyclic Peptide Structure Prediction}
 
\author[1]{Yongchan Hong\thanks{Equal contribution.}}
\author[2]{Zihang Fang\textsuperscript{*}}
\author[2]{Mukta Ubale\textsuperscript{*}}
\author[3]{Yan Liu}
\author[2]{Xingyao Chen\textsuperscript{\textdagger}}
\author[1]{Vsevolod Katritch\thanks{Correspondence to \texttt{xchen@emila.org} and \texttt{katritch@usc.edu}.}}
\affil[1]{Department of Quantitative and Computational Biology, University of Southern California, Los Angeles, CA, USA}
\affil[2]{Ellison Medical Institute, Los Angeles, CA, USA}
\affil[3]{Department of Computer Science, University of Southern California, Los Angeles, CA, USA}
\affil[ ]{\texttt{hongyong@usc.edu}, \texttt{\{zfang,mubale,xchen\}@emila.org}, \texttt{yanliu.cs@usc.edu}, \texttt{katritch@usc.edu}}
\date{}

\begin{document}

\maketitle

\begin{abstract}
Macrocyclic peptides have drawn growing attention as a therapeutic modality, and co-folding models are increasingly applied to predicting their structures. Yet published evaluations rely on small, loosely curated target sets that differ from paper to paper, so reported results are not comparable, and most report accuracy through pose RMSD alone, without checking whether the predicted molecule is chemically correct. We introduce PRETZEL (Prediction of Ring-Enclosed Topology: Zero-shot Evaluation of cycLic peptides), a benchmark of 285 curated cyclic peptide–protein complexes spanning three cyclization chemistries, all post-dating the training cutoff of every model evaluated. PRETZEL comprises three components: PRETZEL-SET, an evaluation set annotated by cyclization chemistry and chemical novelty; PRETZEL-TWIST, a specification layer that compiles a single canonical macrocycle description into each engine's native input format and verifies the returned structure; and PRETZEL-SALT, an evaluation suite pairing standard pose metrics with chemical validity checks for bond distance, bond angles, dihedrals, and ring closure. Using these, we evaluate four engines: AlphaFold3, Boltz-2, Protenix-v2, and HighFold3. Protenix-v2 achieves the best median RMSD at 9.05 Å, with 19.3\% of predictions under 2 Å and 37.5\% under 5 Å. Performance varies by cyclization chemistry: models handle head-to-tail macrocycles relatively well but degrade on stapled targets. Chemical validity is even weaker, with at most 8.1\% of predictions having a fully valid set of inter-residue bonds, and no engine closes every macrocycle in more than 80.7\% of its predictions, which indicates that pose accuracy alone is an insufficient criterion for these models. These failures share a cause: co-folding models place macrocycles without representing them, treating cyclization and non-canonical chemistry as annotations layered onto a residue sequence rather than as first-class structure. PRETZEL is released with all inputs, predictions, and evaluation code.
\end{abstract}

\section{Introduction}
\label{sec:intro}

Macrocyclic peptides occupy a chemical space that neither small molecules nor
antibodies reach \citep{driggers2008macrocycles, villar2014macrocycles}, and
interest in them as therapeutics has grown accordingly
\citep{vinogradov2019macrocyclic, muttenthaler2021trends}. Closing the backbone into a ring removes much of a peptide's conformational freedom, which lowers the entropic cost of binding, protects the amide bonds from exopeptidases, and, for certain ring chemistries, confers enough passive permeability to reach intracellular targets \citep{vinogradov2019macrocyclic, zorzi2017cyclic}. Macrocycles also cover the extended, shallow surfaces that mediate
protein--protein interactions, which small molecules bind poorly and antibodies
cannot reach at all. Generative methods such as RFpeptides
\citep{rettie2025rfpeptides} now produce macrocyclic binders de novo, and each
of such campaigns depends on the ability to predict where the designed peptide sits on
its target.

All-atom co-folding models are the tools most often applied to this problem.
AlphaFold3 \citep{abramson2024alphafold3} and various foundational models  \citep{wohlwend2024boltz1, passaro2025boltz2, bytedance2025protenix,
protenix2026v2} can in principle express a macrocycle through arbitrary chemistry and explicit covalent bonds, and a second line of work targets cyclization directly by modifying the positional encoding \citep{rettie2025afcycdesign,
zhang2024highfold, zhang2025highfold3, xie2025cyclicboltz1, mao2025ncpepfold,
vilya2026vilya2}. What has not kept pace is how they are evaluated. Existing peptide resources \citep{zhang2026pepbenchmark, wen2019pepbdb, hauser2016leadspep,
zhai2025peppcbench} are not organized around macrocycles: they treat peptides as a single class without stratifying by cyclization chemistry, and their metrics do not test the specific chemistry of ring closure or stereochemistry. Most importantly, they often do not account for model training cutoff dates, and thus raise a serious risk of data leakage.


The absence of a macrocycle-specific benchmark shows up in three ways. First, published evaluations use small, non-overlapping target sets. HighFold3 \citep{zhang2025highfold3} is assessed on 63 monomers inherited from AfCycDesign \citep{rettie2025afcycdesign} and 17 internal complex dataset, while other models like NCPepFold use their own hand curated 31 cyclic peptides \citep{mao2025ncpepfold}. Few report whether the structures postdate the training cutoffs of the models compared, so an apparent success may reflect recall rather than prediction. Second, what is supplied to each engine is neither documented nor equivalent across engines: a head-to-tail macrocycle is a cyclization flag to Boltz-2, an explicit bond record to Protenix, and a bond AlphaFold3 removes before inference. Papers report which models they ran but not
how the macrocycle was expressed, so part of what is being compared is the configuration procedure. Third, accuracy is reported through pose metrics alone, which are insensitive to what the coordinates represent. A prediction can fall within 2 Å of the reference while leaving the macrocycle open, stretching or collapsing the bonds that close it, or distorting the geometry at the very junctions that distinguish a macrocycle from a linear peptide. For a modality defined by its ring closure and by chemistry that sits outside the standard residue alphabet, this is a substantial gap in what is being measured.



\begin{figure}[h]
\begin{center}
\includegraphics[width=\linewidth]{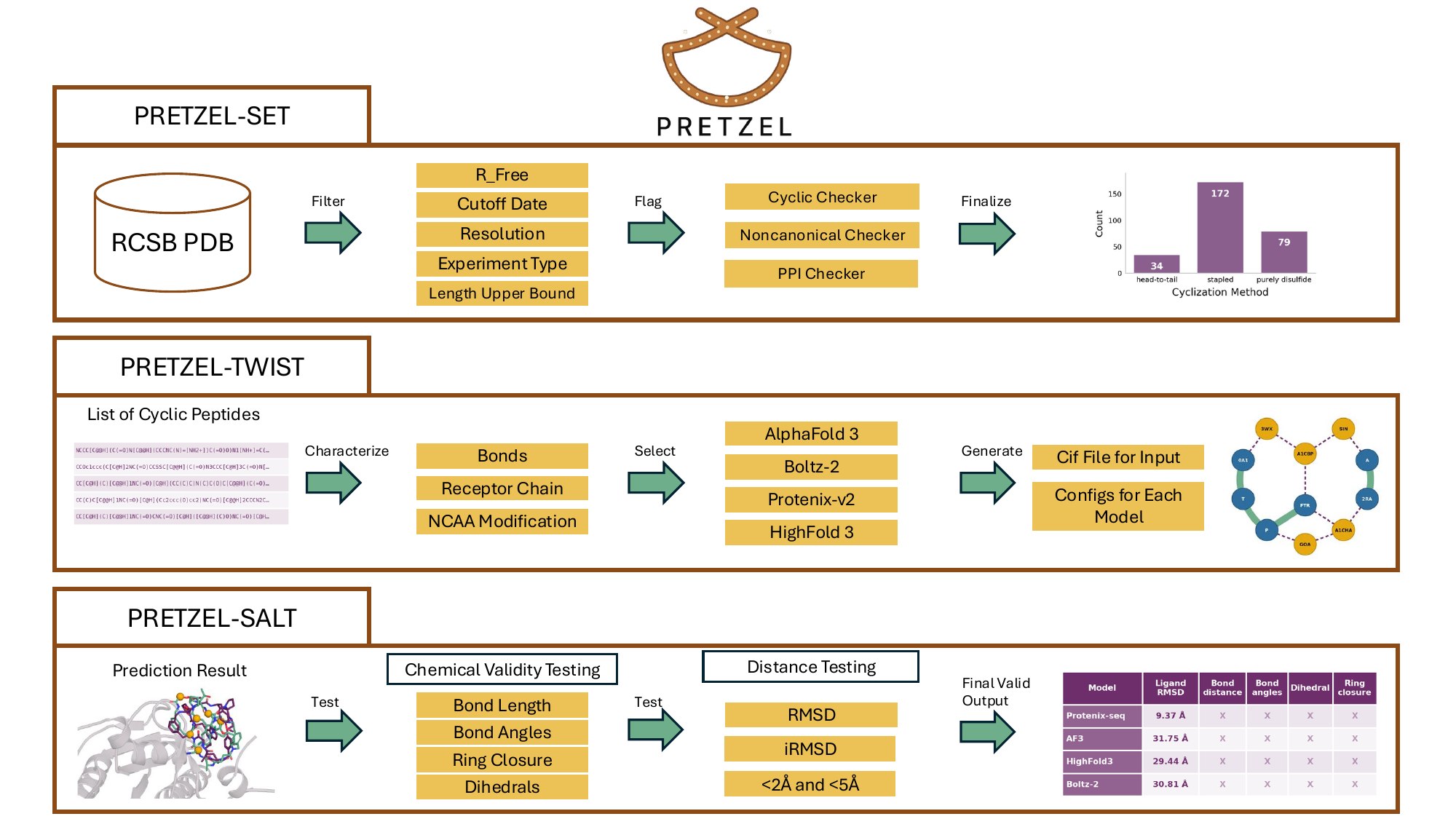}
\end{center}
\caption{Architecture of PRETZEL. PRETZEL-SET filters and flags RCSB PDB entries into 285 cyclic peptide--protein complexes stratified by cyclization chemistry. PRETZEL-TWIST compiles each complex from a single canonical description into the native input format of every evaluated engine. Here PDB 10SB compiles into two sequence segments (blue, backbone in green) plus five ligand entities (gold) for the parts no sequence can carry, stitched together by nine covalent bonds the input must declare outright (dashed). PRETZEL-SALT scores predictions on
chemical validity as well as pose accuracy.}
\label{fig:PRETZEL}
\end{figure}

We introduce PRETZEL (Prediction of Ring-Enclosed Topology: Zero-shot Evaluation
of cycLic peptides), a benchmark of 285 curated
cyclic peptide--protein complexes, all postdating the training cutoff of every model evaluated (2023-06-01). It is composed of three components (Figure \ref{fig:PRETZEL}):

\begin{itemize}
  \item \textbf{PRETZEL-SET.} 285 complexes selected from the PDB under a filter we release in full, annotated by cyclization chemistry (head-to-tail amide, head-to-tail disulfide, inner disulfide, side-chain
    staple) and by chemical novelty, so that accuracy can be read per cyclization type
    rather than averaged into a single figure.
  \item \textbf{PRETZEL-TWIST.} A specification layer that reads each macrocycle once into a canonical, engine-independent description covering entity layout, every inter-residue bond with both endpoints typed, non-canonical residues and receptor contacts, then compiles that description into each engine's native input format and verifies that the returned structure is the molecule requested. This makes the input side of the comparison auditable.
  \item \textbf{PRETZEL-SALT.} An evaluation suite that pairs standard pose accuracy (RMSD) metrics with chemical validity checks for bond distance, bond angles, dihedrals and ring closure.
\end{itemize}

We use these to evaluate four models: AlphaFold3
\citep{abramson2024alphafold3}, Boltz-2 \citep{passaro2025boltz2}, Protenix-v2  \citep{protenix2026v2} and HighFold3 \citep{zhang2025highfold3}. Models tend to show poor results with Protenix-v2 obtaining the best median RMSD at $9.05$\,\AA{}, with $19.3\%$ of predictions below 2\,\AA{} and $37.5\%$ of predictions below 5\,\AA{}. Chemical validity is worse across the predictions, with at most 8.1\% pass the bond-distance validity check and also showing less than half of the predictions passing bond angles, dihedral and ring closure predictions. Taken together these point to a computational challenge: co-folding models place macrocycles
without representing them: cyclization is supplied through channels that leave no measurable
trace in the output, and the result is scored by metrics and confidence heads that does not check
whether the predicted object is the requested molecule. We have released the code on \href{https://github.com/yongchand/PRETZEL}{Github}.


\section{Related Work and Prior Knowledge}
\label{sec:related}

\paragraph{Cyclic peptides.}

A cyclic peptide is a peptide where one or more covalent bonds close a ring through the backbone, the side chain, or both.
Rings close through several distinct chemistries \citep{white2011macrocyclization}: a head-to-tail amide joining the N- and
C-termini, a lactam linking a side chain to a terminus or to a second side chain, a disulfide bridge between two cysteines, a thioether, and side-chain staples that close a ring through a synthetic linker
rather than a backbone amide \citep{lau2015stapling}.
Therapeutic macrocycles routinely combine these
closures with D-amino acids, $N$-methylated backbone amides and other
non-canonical residues, which underlie their protease resistance and membrane
permeability \citep{vinogradov2019macrocyclic, zorzi2017cyclic}.

\paragraph{Structure prediction models.}
AlphaFold3 \citep{abramson2024alphafold3} introduced an all-atom diffusion
decoder able to represent arbitrary chemistry through the Chemical Component
Dictionary and to accept declared covalent bonds, and open reproductions
followed with Boltz-1 \citep{wohlwend2024boltz1}, Boltz-2
\citep{passaro2025boltz2}, and Protenix \citep{bytedance2025protenix,
protenix2026v2}. Since their main objective was not for cyclic peptide, their handling of ring closure differs: AlphaFold3 discards polymer--polymer bonds before inference, Protenix admits them only for cyclic peptides, and Boltz-2 alone encodes cyclicity in the model, via a per-chain flag.

A separate line of work adds cyclization explicitly. AfCycDesign
\citep{rettie2025afcycdesign} showed that a cyclic offset applied to AlphaFold2's
relative positional encoding is sufficient to enforce a ring, and HighFold
\citep{zhang2024highfold} extended the idea to complexes with head-to-tail and
disulfide constraints. Later work targets non-canonical residues: HighFold3 \citep{zhang2025highfold3} carries the offset onto the
AlphaFold3 framework, CyclicBoltz1 \citep{xie2025cyclicboltz1} adapts the offset
to Boltz-style tokenization so that atom-wise non-canonical residues are
still treated as single residues, and NCPepFold \citep{mao2025ncpepfold} combines
a cyclic position matrix with a multi-granular residue and atom representation. A detailed explanation will be given in Appendix \ref{app:landscape}.


\paragraph{Benchmarks.}
\begin{table}[t]
\centering
\caption{Existing benchmarks relevant to cyclic peptide structure prediction. ``n'' is the total number of macrocycles in the benchmark dataset.
``Post-cut.'' is the number of targets deposited after 2023-06-01, the training cutoff of
the most recently trained engine evaluated in the paper (Boltz-2).}
\label{tab:related-benchmarks}
\vspace{0.5em}
\small
\setlength{\tabcolsep}{6pt}
\begin{tabular}{@{}l l r r@{}}
\hline
Benchmark & Content & $n$ & Post-cut. \\
\hline
\multicolumn{4}{@{}l}{\textit{Cyclic peptide monomers}} \\
AfCycDesign \citep{rettie2025afcycdesign} & NMR monomers, 4--39 res. & 80 & 0 \\
HighFold monomer \citep{zhang2024highfold} & Disulfide-bearing subset & 63 & 0 \\
\multicolumn{4}{@{}l}{\textit{Cyclic peptide--protein complexes}} \\
ADCP \citep{zhang2019adcp} & Backbone (18) + SS (20) & 38 & 0 \\
HADDOCK2.4 \citep{charitou2022haddock} & Backbone (18) + SS (12) & 30 & 0 \\
HPEPDOCK2.0 \citep{tao2022hpepdock} & Disulfide-cyclised, 5--30 res. & 25 & 0 \\
CPSet \citep{zhao2024cpset} & 127 canonical + 366 non-canonical  & 493 & 0 \\
\multicolumn{4}{@{}l}{\textit{General peptide sets}} \\
Riptides \citep{vilya2026vilya2} & Canonical + non-canonical complexes & 65 & 65 \\
\hline
\textbf{\textsc{Pretzel-Set}} & Macrocycle complexes & 285 & 285 \\
\hline
\end{tabular}
\end{table}

Benchmarks have repeatedly redirected how progress in small-molecule structure prediction is measured. PoseBusters
\citep{buttenschoen2024posebusters} is the closest analogue to the present work:
it demonstrated that docking predictions scoring well under RMSD are frequently
physically or chemically implausible, and that evaluating geometry alongside
accuracy changes the relative ranking of methods. For peptides, comparable
infrastructure remains limited.
PepBenchmark \citep{zhang2026pepbenchmark} standardises datasets, preprocessing
and metrics for peptide property prediction, but operates at the sequence level
and does not address structure. PepPCBench \citep{zhai2025peppcbench} evaluates
protein--peptide complex prediction across folding networks, and Riptides
\citep{vilya2026vilya2} provides a curated, chemically diverse protein--peptide
set released alongside Vilya-2. The problems compound for  macrocycles: peptides
are treated as a single class rather than stratified by ring chemistry, and the
reported metrics assess pose accuracy without testing ring closure or stereochemistry. Table~\ref{tab:related-benchmarks} lists the existing resources that contain macrocycles. Each was assembled for a single method or docking program, and none is large and diverse enough to perform benchmarking. Every set but Riptides predates 2023 and falls entirely within the training windows of the engines evaluated here; Riptides postdates them but contains 65 macrocyclic targets and does not separate macrocycles from linear peptides.


\section{PRETZEL-SET}
\label{sec:set}

We constructed \textsc{Pretzel-Set} using a systematic two-stage pipeline to curate high-quality cyclic peptide structures from the RCSB Protein Data Bank \citep{berman2000pdb, burley2026rcsb}.

Using the RCSB PDB API query tool, we retrieved all entries released after 2023-06-01, the training cutoff of Boltz-2 \citep{passaro2025boltz2} and the latest among the models evaluated. We defined peptides as protein polymer entities with sequence lengths of
$\leq 50$ residues. We applied resolution ($\leq 3.0$\,\AA{}) and $R_{\text{free}}$
($\leq 0.5$) cutoffs to exclude poorly modeled structures. 
We included all major crystallographic and
cryo-EM methods. As of September 10th 2026, this query returned 3,787 entries containing protein polymers meeting these criteria. This initial set includes both linear and cyclic peptides, along with potential duplicates.

In the second stage, we performed detailed structural characterization to identify genuine cyclic peptides. Our cyclization detection algorithm constructs a residue graph for each peptide chain, analyzing all inter-residue connections from the CIF structure file. Additionally we applied comprehensive structural quality checks to remove entries that did not meet our standards for reliable benchmarking (Appendix \ref{sec:dataset}). These cumulative filters reduced the 3,787 initial entries to 280 passing complexes suitable for benchmarking. Since each complex may contain multiple peptide entities, we resolved each complex into individual peptide chains and retained one representation per redundant chain group. As a result, we had a total of 285 peptide chains drawn from 280 PDB entries. We deliberately did not apply two filters: (1) we did not exclude peptides chemically similar or identical to molecules already in the pre-cutoff RCSB database; (2) we did not impose any requirements on each peptide's binding protein. Both decisions make our benchmark an unbiased sample of the post-cutoff period, and both make it necessary to describe what the result set contains.

\begin{figure}[h]
\begin{center}
\includegraphics[width=0.8\linewidth]{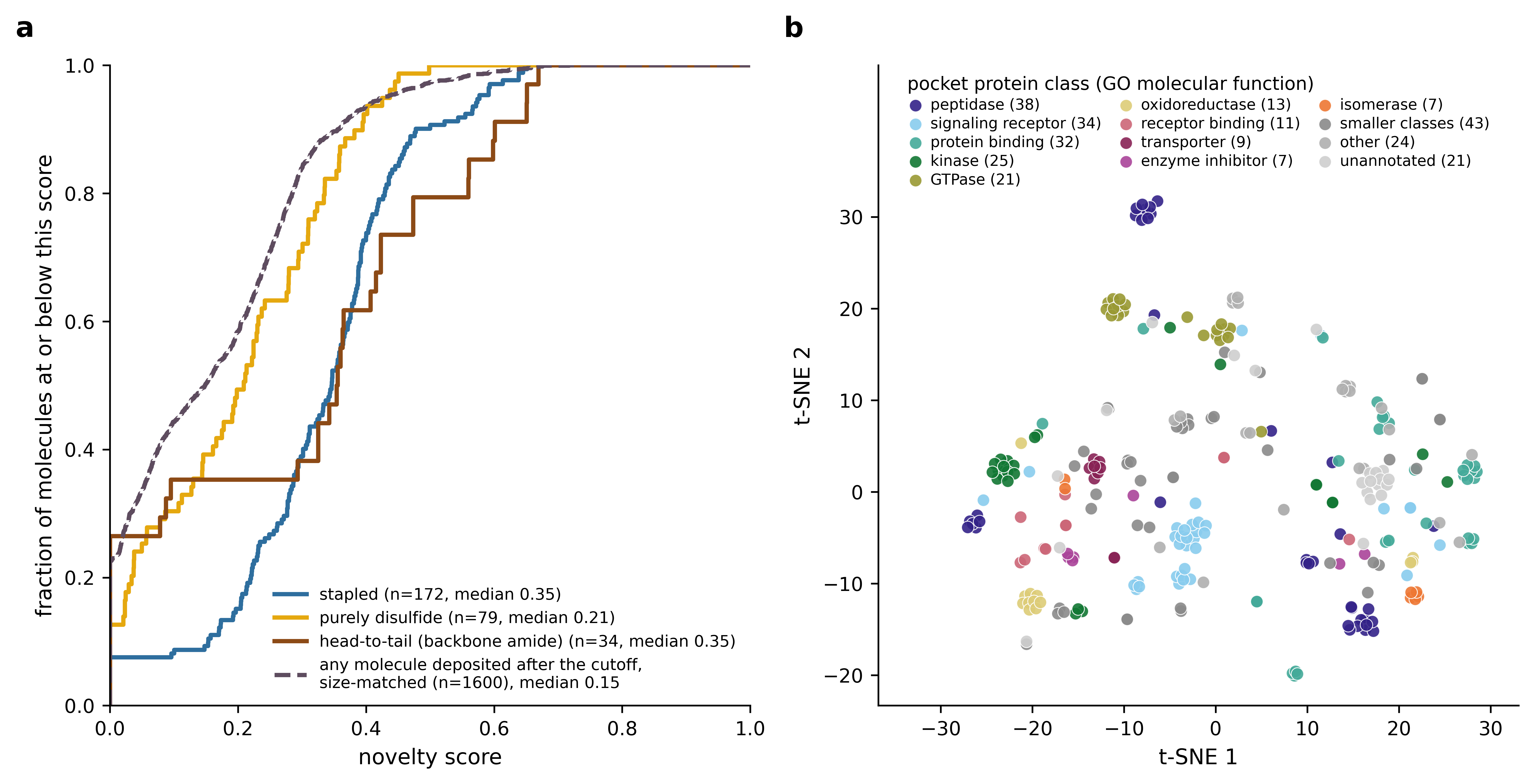}
\end{center}
\caption{Chemical novelty and binding-site diversity of \textsc{Pretzel-Set}.
\textbf{(a)} Cumulative novelty-score distribution for the 285 peptide chains by
cyclization chemistry: head-to-tail, stapled, and purely disulfide against a size-matched sample of all post-cutoff
depositions.
\textbf{(b)} t-SNE of receptor binding pockets colored by GO molecular function}
\label{fig:characterization}
\end{figure}

In order to better characterize the dataset, we have analyzed chemical novelty and binding-site diversity of PRETZEL-SET (Figure \ref{fig:characterization}).

For chemical novelty, we compared each peptide against every molecule released before the
cutoff using count-based ECFP4 fingerprints \citep{rogers2010ecfp} and MinMax Tanimoto
similarity \citep{swamidass2005minmax}, and define its novelty score as one minus the
closest match. 
The benchmark is appreciably further from the
models' training chemistry than an arbitrary structure from the same window
(Figure~\ref{fig:characterization}a), though the three cyclization classes are not equally
novel: head-to-tail macrocycles are the most distant from prior chemistry and purely
disulfide the least.

For receptors, we defined each pocket as the residues of the dominant receptor chain within
10\,\AA{} of the peptide, following the definition used by PXMeter \citep{chen2025pxmeter}
for pocket-aligned RMSD so that characterization and scoring agree. 
Embedding $1-\texttt{pocket\_qcov}$ (\citep{durairaj2024plinder}) with t-SNE (Figure~\ref{fig:characterization}b) places sites with similar pockets near one another, colored by the receptor's GO molecular-function class (\citep{go2026,ashburner2000go}). The 285 sites span 19 distinct classes, the ten most common drawn individually and the remainder pooled, with the tighter groups being single targets solved repeatedly or close paralogues. Because the majority of site pairs do not share pocket residues, the panel shows both structural and functional diversity. Detailed implementation and dataset analysis results are given in Appendix \ref{sec:set-analysis}.

\section{PRETZEL-TWIST}
\label{sec:twist}

After constructing the final PRETZEL-SET, we developed the PRETZEL-TWIST layer to perform comprehensive characterization of each passing cyclic peptide. We extracted all chemical and structural information into a central model-independent manifest, which served as input into engine-specific  config builder scripts that translated the characterization into the respective formats.

For each peptide entry, we parsed the mmCIF coordinates using residue-to-residue connectivity to determine which residues could be chained together in a single polymer entity and where the chain had to split. Our criteria followed the universal linking rule: consecutive protein residues bond through their backbone C and N atoms. Non-canonical amino acids (NCAAs) with complete backbone atoms were labeled as modifications and NCAAs with missing backbone atoms or synthetic fragments were flagged as ligand entities. Ligand entities that carry no covalent bonds to the peptide, such as detached caps or floating solvent, were automatically dropped. Terminal unmodelled residues were trimmed. 

Every inter-residue covalent bond was classified across two dimensions: (1) the molecule types at both ends (protein or ligand), and (2) whether the bond was implicit (free with the polymer declaration, e.g., standard backbone amides) or explicit (requiring model-specific declaration). Receptor chains were selected using a contact rule: any polymer chain within 4.5\AA\ of the peptide, excluding copies of the peptide entity itself. Each peptide chain receives its own contact list; configurations declared all binding partners in the interaction radius. Bonds were subtyped as backbone, head-to-tail (N' → C' cyclization), head-to-tail disulfide, inner disulfide, side-chain staple, or linker, allowing builders to apply model-appropriate handling strategies (Appendix \ref{app:definition}). 

For this benchmark study, we ran the PRETZEL-SET through four cofolding models: AlphaFold3, HighFold3, Boltz-2, Protenix-v2. Each builder performed three steps: (1) translating segment and ligand entity definitions into the engine's chain-naming and chain-numbering scheme; (2) declaring all explicit bonds according to the engine's syntax and limitations; (3) attaching pre-computed multiple sequence alignments for receptor chains. The format and notations followed the examples and descriptions provided by each model (Appendix \ref{sec:engine-examples}). All four engines share an AF3-style architecture, combining residue- and atom-level representations with diffusion-based coordinate generation, but differ in chemical featurization and positional encoding. AF3 pairs atom features, reference conformers and explicit bond-connectivity features with a linear relative positional encoding. HighFold3 replaces those linear offsets within cyclic peptides with topology-aware offsets, so residues across the cyclization junction become sequence neighbours. Boltz-2 keeps standard residues as single tokens while expanding ligands and non-canonical residues into per-atom tokens, adding bond-type features and a cyclic positional encoding. Protenix-v2 widens the AF3-derived network's representation dimensionality and capacity while retaining explicit covalent-connectivity conditioning. Detailed configurations for each model are explained in the Appendix \ref{sec:engine-examples}.



\section{PRETZEL-SALT}
\label{sec:salt}
After running each cofolding model, we evaluate the predictions using PRETZEL-SALT, which comprises both structural accuracy metrics and chemical validity metrics.  For peptide pose evaluation, we implemented an extension of the pocket-aligned RMSD approach of PXMeter (Ma et al., 2025) to peptides distributed across multiple chains.
Matched model pocket atoms are superposed onto the reference pocket by least-squares rigid-body fitting, and this transformation is applied to the model peptide with no additional fitting.
The reported ligand RMSD is the distance between matched reference and transformed model peptide-entity atoms, including backbone and side-chain atoms. Residues are matched by name and covalent connectivity, and atoms by name within matched residues; unmatched atoms are excluded. Among enumerated equivalent residue mappings, the lowest pocket-aligned ligand RMSD is reported. We have also calculated secondary metrics iRMSD and backbone RMSD (Appendix \ref{app:accuracy})

Beyond accuracy, PRETZEL-SALT assesses chemical validity at two levels. The standard level
evaluates inter-residue bond distance; intra-residue bonds are excluded because their geometry
comes from the Chemical Component Dictionary (CCD) rather than being predicted. Every inter-residue
bond is perceived once, in the deposited structure --- from annotated links, chain
connectivity, and coordinates for crosslinks the deposition leaves undeclared --- and located
in the prediction by the same two atoms. No bond perception is performed on predicted
coordinates, so a bond stretched past recognition is still measured rather than silently
dropped. A bond passes when its length falls inside a type-specific reference band or
reproduces the deposited value to within 0.05 \AA.

For the advanced validity test, we check bond angles, dihedrals and ring closure. At
the standard level, every measurement is perceived from the deposition and looked
for in the prediction, and passes if it falls inside its reference band or
reproduces the deposited value. Bond angles are measured at every inter-residue
junction and scored against a named restraint where one exists, otherwise against
the vertex atom's hybridization. Dihedrals pool three classes --- $\omega$, carbonyl planarity and the CA improper magnitude ---
each scored against the nearest of its permitted targets.  Ring closure requires every bond of each macrocycle perceived in the deposition to be intact in the
prediction; perceiving rings in the prediction instead would let a model that never closed the ring have nothing to report, so the failure would pass silently. Full chemical validity details are given in Appendix \ref{sec:chem-validity}.


\section{Experiments}
\label{sec:experiments}
We organize our empirical analysis around four research questions:
\textbf{(RQ1)} Do the engines predict accurate poses on \textsc{Pretzel-Set}?
\textbf{(RQ2)} Does performance depend on cyclization chemistry?
\textbf{(RQ3)} Does performance depend on chain configuration type?
\textbf{(RQ4)} Are the predicted macrocycles chemically valid?

\subsection{Experimental Setup}
\label{sec:setup}
We evaluate four all-atom co-folding engines, AlphaFold3, Protenix-v2, Boltz-2 and HighFold3, each run on configurations generated by \textsc{Pretzel-Twist}. Engines run at default settings, except that we fix the sample count at five for all of them, since defaults differ and sampling budget would otherwise confound the comparison (Additional Experiment in Appendix \ref{sec:additional-experiment}). For each target we score the structure the engine ranks first by its own confidence, with no reference to the deposited structure. For pose accuracy we report mean and median RMSD together with success rates at 2\AA{} and 5\AA{}, following the accuracy scale conventional in protein-peptide docking, where sub-2\AA{} models are considered near-native and 2--5\AA{} models of usable quality \citep{ciemny2018pepdock, raveh2011flexpepdock}. For chemical validity we report both levels, covering bond distance at the standard level and bond angles, dihedrals and ring closure at the advanced level. Per-engine versions, seeds and remaining sampling settings are given in Appendix \ref{app:reproducibility}. Before beginning the experiments, we ran a simple baseline test on the dataset that HighFold3 was tested on to show that our experiment is trustworthy (Appendix \ref{sec:baseline}).

\added{Because all four engines are scored on the same 285 peptide chains
($1{,}140$ predictions in total), their scores are not independent, and the groups
we compare also differ in peptide length, non-canonical residue content and
covalent complexity. To check that our comparisons hold once this is accounted
for, we additionally fit a mixed-effects model of $\log$ RMSD with a random
intercept per peptide and these properties as covariates, and we test whether
engines respond differently to cyclization class and chain configuration using
within-peptide $F$ tests. The model and full results are given in
Appendix~\ref{sec:statistical-analysis}.}



\subsection{Experimental Results}
\label{sec:results}

\begin{table}[t]
\centering
\caption{Structural accuracy and chemical validity of four co-folding models on 285 peptide
targets. $N$ is the number of predicted structures returned by each
model; all percentages use that model's $N$ as the denominator. Higher is better except for
median and mean RMSD. *For one target, Boltz ran with a single diffusion sample due to an out-of-memory (OOM) error.}
\label{tab:accuracy-validity}
\vspace{0.5em}
\small
\setlength{\tabcolsep}{3.5pt}
\begin{tabular}{@{}l cc cc cc c ccc@{}}
\hline
& & & \multicolumn{2}{c}{\makecell{RMSD (\AA)}}
& \multicolumn{2}{c}{\makecell{Success\\(\% of $N$)}}
& \multicolumn{4}{c}{Chemical validity (\% of $N$ passing)} \\
\cline{4-5}\cline{6-7}\cline{8-11}
& & & & & & & Standard & \multicolumn{3}{c}{Advanced} \\
\cline{8-8}\cline{9-11}
Model & \makecell{Training\\cutoff} & $N$ & Median & Mean & $<2$\,\AA & $<5$\,\AA
& \makecell{Bond\\distance} & \makecell{Bond\\angles} & Dihedral & \makecell{Ring\\closure} \\
\hline
  AlphaFold3  & 2021-09-30 & 285 & 10.04 & 15.48 & 17.5 & 31.2
              & \textbf{8.1} & 14.4 & 43.2 & 66.0 \\
  HighFold3   & 2021-09-30 & 285 & 9.80 & 15.06 & 18.9 & 32.6
              & \textbf{8.1} & 14.4 & 46.0 & 66.7 \\
  Boltz-2     & 2023-06-01 & 285* & 9.79 & 16.29 & 14.7 & 34.4
              & 4.6 & 6.3 & 39.3 & 58.2 \\
  Protenix-v2 & 2021-09-30 & 285 & \textbf{9.05} & \textbf{14.52} & \textbf{19.3} & \textbf{37.5}
              & 4.2 & \textbf{16.8} & \textbf{53.0} & \textbf{80.7} \\
\hline
\end{tabular}
\end{table}


\subsubsection{RQ1: Prediction Accuracy}
Table~\ref{tab:accuracy-validity} shows that all four engines struggle on \textsc{Pretzel-Set}: median RMSD ranges from 9.05 to 10.04\,\AA, and no model places more than a fifth of its predictions within 2\,\AA{} of the deposited structure. Protenix-v2 slightly leads on every accuracy measure, with a median of 9.05\,\AA{} and 55 of its 285 returned structures under 2\,\AA{} (19.3\%) and 37.5\% of its predictions at the looser 5\,\AA{} threshold. HighFold3's cyclization-aware positional encoding yields a slight gain over its own AlphaFold3 backbone, but not enough to overtake a general-purpose model, suggesting that explicit ring handling helps without being sufficient on its own. A detailed case study is included in Appendix \ref{sec:case-study}.

\subsubsection{RQ2: Cyclization Chemistry}
\begin{figure}[h]
\begin{center}
\includegraphics[width=\linewidth]{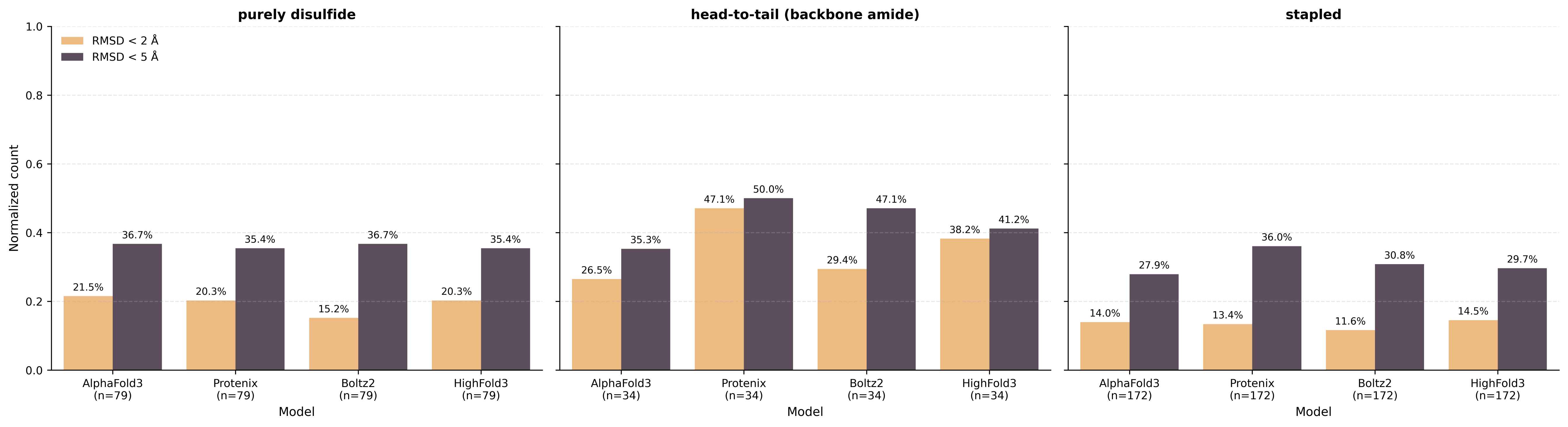}
\end{center}
\caption{Normalized design success of each model and cyclization group based on 2Å and 5Å cutoff.}
\label{fig:cyclization}
\end{figure}

Stratifying by cyclization type (Figure~\ref{fig:cyclization}) shows that difficulty
tracks the chemistry of the macrocyclic linkage\added{, though the engine ordering is not fixed: AlphaFold3
is weakest on head-to-tail targets and strongest on disulfide-only ones. The shift
does not survive multiple-testing correction ($p=0.017$ raw, $0.10$ after Holm), so
we claim no engine-specific effect}. Head-to-tail cyclized targets are the most
tractable for every engine (26.5--47.1\% under 2\,\AA\ and 35.3--50.0\% under
5\,\AA), stapled peptides the hardest (11.6--14.5\% under 2\,\AA\ and 27.9--36.0\%
under 5\,\AA). HighFold3 gains slightly over AlphaFold3 on head-to-tail targets,
suggesting that the benefit of an explicit cyclization prior does not extend beyond
the linkage types it encodes\added{; Protenix-v2, however, leads this class outright
(47.1\% under 2\,\AA) without any such prior}. Explicit disulfide bonds could not be
modeled by HighFold3 in cases with ligands or NCAAs (Appendix~\ref{sec:limitation}),
so our predictions only have implicitly-set disulfide bonds, explaining the
similarity in RMSD values for these two models.
\added{Synthetically stapled macrocycles are the hardest case for every model tested.
The classes are not matched on other properties, however: head-to-tail targets
average 10.3 residues against 23.6 for disulfide, and stapled targets carry the most
cross-links (3.0 against 1.7). Adjusting for length, non-canonical content and
covalent complexity, the stapled--head-to-tail gap narrows from $+88\%$ to $+42\%$
and is no longer significant ($p=0.11$), while cross-link count remains strongly
predictive ($+46\%$ per cross-link, $p=0.0008$). Cyclization class is better read as
a marker of covalent constraint burden than as an independent determinant of
accuracy.}



\subsubsection{RQ3: Chain Configuration Type}
\begin{figure}[h]
\begin{center}
\includegraphics[width=0.7\linewidth]{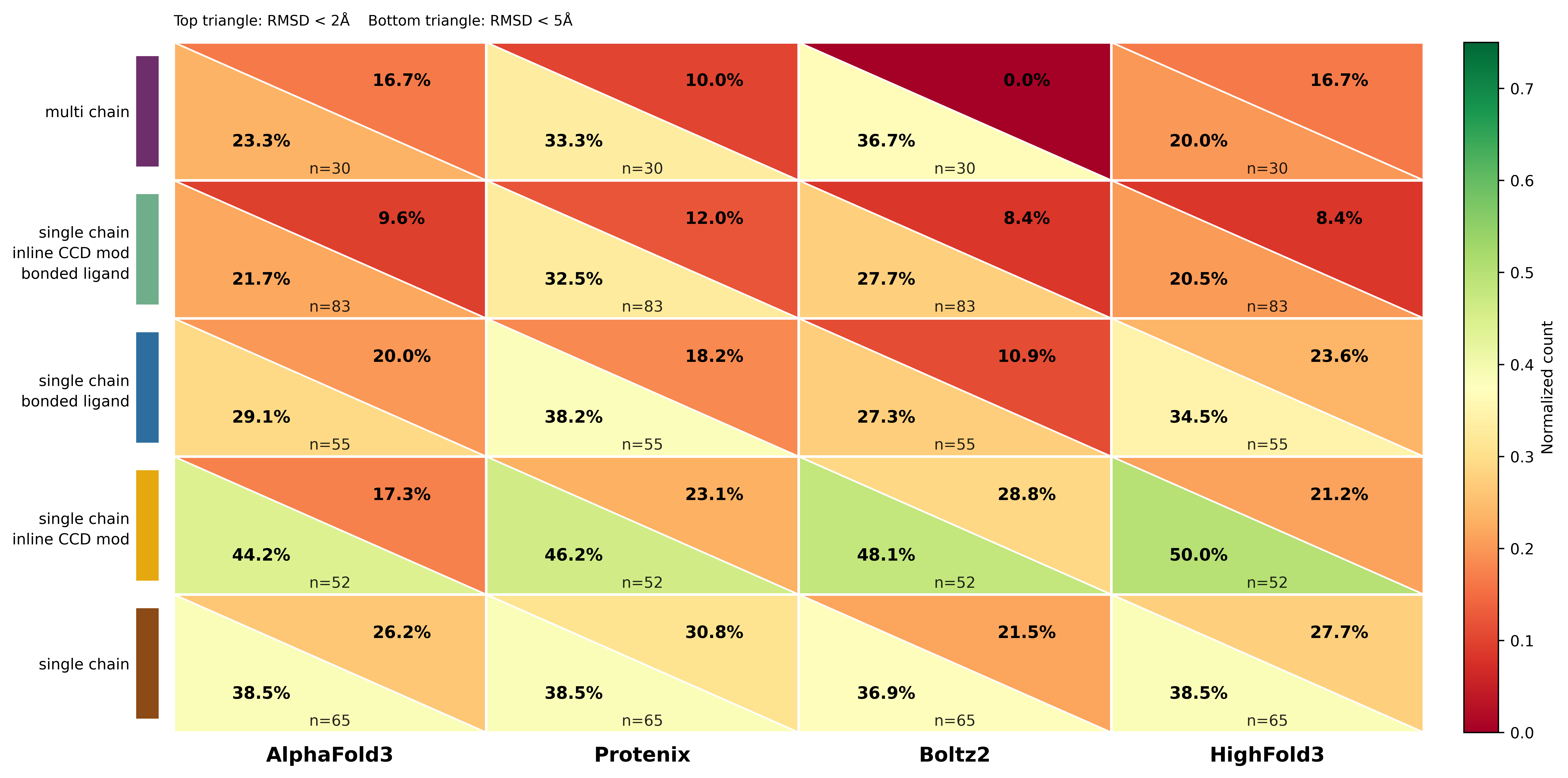}
\end{center}
\caption{Normalized design success of each model and chain configuration type based on 2Å and 5Å cutoff.}
\label{fig:chain}
\end{figure}

Configuration families (Figure~\ref{fig:chain}) are defined not by chemistry but by how a
macrocycle must be expressed to the engine: as a plain sequence, with inline CCD
modifications, with a covalently bonded ligand entity, or as several entities. Plain single
chains and single chains with inline CCD modifications are the easiest families for every
engine (17.3--28.8\% under 2\,\AA, 38.5--50.0\% under 5\,\AA). Single chains carrying both an
inline CCD modification and a bonded ligand are the hardest family in the benchmark, falling to
8.4--12.0\% under 2\,\AA\ and 20.5--32.5\% under 5\,\AA, roughly a third of the plain
single-chain rate at 2\,\AA\ for every engine.
\added{The two features that define that family---an inline CCD modification and a
bonded ligand---act independently rather than compounding. They form a $2\times2$
design across 255 single-chain peptides, and their interaction is negligible both on
RMSD ($p=0.38$ unadjusted, $0.71$ after adjusting for length, non-canonical residue
count, cyclization class and cross-link count) and on the rate under
$2\,\text{\AA}$ ($p=0.36$ and $0.88$). The family is hardest because it carries both
penalties, not because they combine into something worse. The families track the
number of covalent constraints an engine must satisfy at once: cross-link count is
the strongest single chemical predictor of accuracy, at $+46\%$ RMSD per additional
cross-link ($p=0.0008$), and once it is accounted for the gap between stapled and
head-to-tail macrocycles narrows from $+75\%$ to $+42\%$ ($p=0.11$). Configuration
family is what that burden looks like in the input file.}

No engine leads every family: Protenix-v2 is strongest on the hardest family at both
thresholds, while HighFold3 leads the inline CCD family at 5\,\AA. The ordering
across families is otherwise consistent between engines \added{(the
engine~$\times$~family interaction is not significant, $F(12,840)=1.14$, $p=0.32$;
pairwise Spearman correlation over family means 0.9--1.0)}, so difficulty tracks the
representation the input format forces rather than any one engine's handling of it.

\subsubsection{RQ4: Chemical Validity}

\begin{table}[t]
\centering
\caption{Backbone amide geometry by junction type, over 4,246 inter-residue amides from 285
chains. A junction is classified by whether its two flanking residues are canonical (cAA) or
non-canonical (ncAA). Values are the median absolute deviation between predicted and deposited
bond length; lower is better.}
\label{tab:amide}
\vspace{0.5em}
\small
\setlength{\tabcolsep}{3.5pt}
\begin{tabular}{@{}l c cccc@{}}
\hline
& & \multicolumn{4}{c}{Median $|d_{\mathrm{pred}} - d_{\mathrm{dep}}|$ (\AA)} \\
\cline{3-6}
Junction & $n$ & AlphaFold3 & HighFold3 & Boltz-2 & Protenix-v2 \\
\hline
  cAA--cAA   & 3,253 & \textbf{0.009} & \textbf{0.009} & 0.016 & 0.027 \\
  cAA--ncAA  & 727   & \textbf{0.065} & 0.071 & 0.098 & 0.091 \\
  ncAA--ncAA & 266   & 0.095 & 0.091 & 0.139 & \textbf{0.062} \\
\hline
\end{tabular}
\end{table}

Model predictions are frequently invalid before accuracy is considered at all. Bond distance is
the strictest check and the one every engine fails most often: the best returns a fully valid
set of inter-residue bonds in fewer than one prediction in eleven (8.1\%). Bond angles are barely better (6.3--16.8\%), while dihedrals (39.3--53.0\%) and ring closure (58.2--80.7\%) pass more often, as one would expect of
checks with looser tolerances and fewer measurements per structure. The ordering across engines
differs by category, so no engine is uniformly the most chemically reliable, and a structure that
is close in RMSD routinely contains at least one inter-residue bond that is stretched, collapsed
or fused.

Restricting to backbone amides shows where the bond-distance failures concentrate (Table~\ref{tab:amide}). An amide between two non-canonical residues is chemically identical to
one between two canonical residues, with the same atoms and the same equilibrium length, yet AlphaFold3, HighFold3 and Boltz-2 deviate six to eight times further from the deposited distance as soon as one endpoint is non-canonical, having been near-exact on canonical junctions. Protenix-v2 behaves differently: three times worse than the others on cAA--cAA junctions (0.027\,\AA), it degrades least across the three categories and is the most accurate engine when
both endpoints are non-canonical (0.062\,\AA). Read as a whole, a macrocycle is a set of bonds that must simultaneously reach equilibrium, and the two behaviours are different solutions to that problem: three engines hold the bonds they can express well at the cost of the rest, while
Protenix-v2 distributes the error more evenly. Its training-free guidance module, which constrains steric and torsional geometry during sampling rather than relying on the representation alone, is a plausible source of that balance, and is consistent with its leading dihedral and ring-closure rates while ranking last on bond distance overall.

Chemical validity and pose accuracy are correlated: every check separates accurate from inaccurate predictions, most strongly bond angles, where 52\% of the 148 passing structures fall within 2\,\AA\ against 12\% of failing ones (Appendix~\ref{sec:val-accuracy}). Model confidence
does not carry that information. ipTM tracks pose accuracy across all four engines ($r = -0.36$ to $-0.51$) but correlates only weakly with chemical validity, and not
significantly with ring closure for any engine ($r = 0.018$--$0.094$; Appendix~\ref{sec:confidence}). An engine that leaves the macrocycle open is therefore about as likely to report high confidence as one that closes it, so the failure is invisible at the point where a practitioner would act on it.

\section{Conclusion}
\label{sec:conclusion}

PRETZEL establishes a reproducible framework for macrocyclic peptide structure prediction,
resolving fragmentation in target curation, engine configuration, and evaluation. All four
co-folding engines struggle on \textsc{Pretzel-Set}: Protenix-v2 leads with a median RMSD of
9.05\,\AA{} and 19.3\% of predictions within 2\,\AA, and no engine returns a fully valid set of
inter-residue bonds in even one prediction in eleven. Behind these numbers lies a computational
challenge: co-folding models place macrocycles without representing them. HighFold3's
architectural prior helps only on the linkage its offset encodes, and Boltz-2, given a dedicated
cyclization flag, closes fewer macrocycles than AlphaFold3, which is never told the ring exists. Meeting this challenge requires encoding cyclization and non-canonical chemistry as first-class structure rather than as annotations layered onto a residue sequence, and evaluating against chemical validity rather than pose alone.

\section*{Ethics Statement}
This dataset was assembled from the RCSB Protein Data Bank, a publicly accessible, open-access resource, and contains no private or sensitive human data. Its main goal is to provide a demanding yet fair benchmark for cyclic peptides and to encourage better approaches to encoding cyclization. We anticipate meaningful societal benefits from this work and foresee no associated harm.

\bibliography{iclr2027_conference}
\bibliographystyle{iclr2027_conference}

\appendix

\newcommand{\TODO}[1]{\textcolor{red}{\textbf{[TODO: #1]}}}

\section{Experiment Reproducibility}
\label{app:reproducibility}

\paragraph{Availability.}
Code (PRETZEL-SET filters, PRETZEL-TWIST builders, PRETZEL-SALT evaluation) is in \url{https://github.com/yongchand/PRETZEL}. Also, evaluation fork for PXMeter is in \url{https://github.com/yongchand/PXMeter-pretzel}.

Generated engine configs, cached MSAs and all raw predicted structures are at \url{https://huggingface.co/datasets/yongchand/PRETZEL}. 

Structures were
retrieved from the RCSB PDB on 2026-09-10; the query is released with the code.
All tables and figures can be regenerated from the released predictions on CPU
alone.

\paragraph{Engine versions and settings.}
We ran AlphaFold3 v3.0.4 (using their provided open-source Docker image) with model parameters downloaded from https://storage.googleapis.com/alphafold3/af3.bin.zst. We ran HighFold3 (commit 1e6125a) with AlphaFold3 v3.0.1 and the same previously mentioned parameters, Boltz-2 v2.2.1 (commit b1ebfc4), and Protenix-v2 v2.0.0 (commit 4c3hhbe). We generated five diffusion samples per target, seed 1 and defaults otherwise. MSA was produced using each engine's documented pipeline, so that every model runs in the configuration its authors intend. Protenix used its bundled ColabFold-compatible search
(\texttt{scripts/colabfold\_msa.py}, ColabFold v1.6.3) against a ColabFold
database mirrored on our own server, \texttt{uniref30\_2302} and
\texttt{colabfold\_envdb\_202108}, computed once per receptor sequence and
reused wherever that receptor recurs, with alignment depth set to 2{,}048.
AlphaFold3 and HighFold3 were given the same alignments, built by AlphaFold3's
own HMMER-based pipeline against the databases its \texttt{fetch\_databases.sh}
installs: UniRef90 2022\_05, MGnify 2022\_05, UniProt 2021\_04, and PDB seqres
2022\_09\_28 for templates. Boltz-2 used the public ColabFold MSA server, 
which runs MMseqs2 against its own hosted copies of UniRef30 and the ColabFold environmental database 
at a release we did not pin. No peptide chain was given an alignment in any engine.


\paragraph{Hardware.}
All inference ran on NVIDIA A100 GPUs 80GB with AMD EPYC 7542 as CPU. 


\section{Landscape of current cyclization models}
\label{app:landscape}

This appendix surveys the models available for macrocyclic peptide structure
prediction, separates them into general-purpose co-folding engines and
cyclization-aware methods, and states why \textsc{HighFold3} was selected as the
cyclization-aware representative evaluated in the main text.

\subsection{General-purpose co-folding engines}
\label{app:landscape:general}

All-atom co-folding models are not designed for macrocycles, but can in
principle express one, because they represent chemistry through the Chemical
Component Dictionary (CCD) and accept declared covalent bonds. They differ in
how a ring is communicated to the model, and this difference is the reason
\textsc{Pretzel-Twist} is required:

\begin{itemize}
  \item \textbf{AlphaFold3}~\citep{abramson2024alphafold3} introduced the
    all-atom diffusion decoder, arbitrary chemistry via the CCD, and declared
    covalent bonds. 
    
  \item \textbf{Boltz-1}~\citep{wohlwend2024boltz1} and
    \textbf{Boltz-2}~\citep{passaro2025boltz2} are open reproductions of the
    AlphaFold3 framework. Boltz exposes cyclization as a per-entity flag
    (\texttt{cyclic: true}) rather than as a bond.
  \item \textbf{Protenix}~\citep{bytedance2025protenix} and
    \textbf{Protenix-v2} accept an explicit bond
    record for the cyclizing linkage. Protenix-v2 additionally applies a
    training-free guidance module constraining steric and torsional geometry
    during sampling.
\end{itemize}

The same macrocycle is therefore handled differently across engines. Papers report which
engines they ran, but not how the ring was expressed, so part of what is
compared across publications is the configuration procedure rather than the
model.

\subsection{Cyclization-aware methods}
\label{app:landscape:cyclic}

A second line of work modifies the positional encoding so that the ring is
enforced architecturally rather than declared as chemistry.

\paragraph{AfCycDesign~\citep{rettie2025afcycdesign}.}
Showed that applying a cyclic offset to AlphaFold2's relative positional
encoding is sufficient to enforce a head-to-tail ring, without retraining. The
offset replaces the linear index distance $|i-j|$ with the circular distance on
the residue ring. Evaluated on 80 NMR monomers (4--39 residues).

\paragraph{HighFold~\citep{zhang2024highfold}.}
Extended the cyclic offset to complexes and to disulfide-bridge constraints,
on the AlphaFold2 framework. Assessed on a 63 monomer and 17 complex subset.

\paragraph{HighFold2~\citep{zhang2025highfold2}.}
Built on the AlphaFold-Multimer framework, HighFold2 extends the predefined
rigid-group definitions and initial atomic coordinates from natural to
non-canonical amino acids (NCAAs), and parameterizes those residues so that
predicted structures can be relaxed by energy minimization to eliminate
clashes.

\paragraph{CyclicBoltz1~\citep{xie2025cyclicboltz1}.}
Carries the cyclic offset onto the Boltz framework. Its contribution is an extension of the offset to the AlphaFold3 tokenization scheme, so that non-canonical residues tokenized atom-wise are still treated as single residues for the purpose of the cyclic encoding. On a HighFold test set of peptides with ncAAs it outperforms HighFold2 on $13/17$ cases, with an average $\mathrm{RMSD}_{\mathrm{C}\alpha}$ of $1.877$\,\AA{} and an average all-atom
RMSD of $3.361$\,\AA.

\paragraph{NCPepFold~\citep{mao2025ncpepfold}.}
Built on RoseTTAFold All-Atom rather than an AlphaFold backbone, NCPepFold
combines a Cyclic Position Matrix with a multigranular representation that mixes
residue-level and atom-level tokens, splitting each non-canonical residue into a canonical amino acid and a covalently attached side-chain group. Its cyclic position matrix is formulated for head-to-tail linkages, so disulfide and side-chain-stapled macrocycles are outside its scope.

\paragraph{HighFold3~\citep{zhang2025highfold3}.}
Carries the cyclic offset onto the AlphaFold3 framework. Built on the
frozen AlphaFold3 pretrained model, it introduces a Cyclic Position
Offset Encoding Matrix (CycPOEM) together with a ``Cyclization Switch'' module
that dynamically selects a linear or cyclic positional encoding matrix at
inference. It comprises two submodels, HighFold3-Linear and HighFold3-Cyclic.

\paragraph{All-atom alternatives.}
Vilya-1~\citep{vilya2026vilya1} and Vilya-2~\citep{vilya2026vilya2} take a
different route, discarding residue tokens in favour of an all-atom chemical
graph. Neither is publicly available at the time of writing, so neither is evaluated here.

\subsection{Why HighFold3}
\label{app:landscape:why}

We evaluate one cyclization-aware engine alongside three general-purpose
engines. \textsc{HighFold3} was selected for three reasons.

\paragraph{(i) It is the reported state of the art among cyclization-aware
methods.} HighFold3 reports outperforming HighFold, HighFold2, CyclicBoltz1,
NCPepFold, CABS-flex, ESMFold and HelixFold on cyclic peptide structure
prediction. On the 63 cyclic peptide monomers of the AfCycDesign dataset it
reports a median $\mathrm{RMSD}_{\mathrm{C}\alpha}$ of $0.918$\,\AA{} and a mean
of $1.337$\,\AA, against HighFold's $1.058$\,\AA{} and $1.478$\,\AA{}
($13.2\%$ and $9.5\%$ improvements); it achieves
$\mathrm{RMSD}_{\mathrm{C}\alpha} < 1.5$\,\AA{} on $44/63$ samples, against
$37/63$ for HighFold and $38/63$ for CyclicBoltz1. On 34 monomer and complex
structures containing unAAs it reports a median $\mathrm{RMSD}_{\mathrm{C}\alpha}$
of $1.711$\,\AA{} and a mean of $1.907$\,\AA, against HighFold2's $1.891$\,\AA{}
and $2.152$\,\AA. Selecting HighFold3 therefore gives the cyclization-aware
family its strongest published representative.

\paragraph{(ii) It shares a backbone with an engine we already evaluate.}
Because CycPOEM is applied to a frozen AlphaFold3, the HighFold3--AlphaFold3
pair isolates the contribution of the cyclic positional encoding from every
other architectural difference. Any gap between them on
\textsc{Pretzel-Set} is attributable to the encoding rather than to training
data, parameter count, or decoder design.

\paragraph{(iii) It is runnable on the full benchmark.} HighFold3 accepts
complexes and non-canonical residues, which is a prerequisite for
\textsc{Pretzel-Set}. Several cyclization-aware methods are monomer-only,
restricted to head-to-tail chemistry, or not publicly released.

\section{PRETZEL-SET: Dataset implementation and composition}

\label{sec:dataset}
We utilized the rcsb-api Python package to programmatically access the RCSB PDB (\citep{berman2000pdb}). The experimental methods were filtered to X-ray diffraction, electron microscopy, electron crystallography, neutron diffraction, powder diffraction, and solution scattering. Resolution and R-free filters were set in order to exclude complexes that were broken, unrefined, or had failed structures.  

\paragraph{RCSB Query: Find Peptides.}\leavevmode
\begin{lstlisting}[style=cfg]
    EXPERIMENT_TYPES = ["ELECTRON CRYSTALLOGRAPHY", "ELECTRON MICROSCOPY", "EPR", "NEUTRON DIFFRACTION", 
    "POWDER DIFFRACTION", "SOLUTION SCATTERING", "X-RAY DIFFRACTION"]
    RESOLUTION = 3
    R_FREE = 0.5
    LENGTH_UPPER_BOUND = 50
    CUTOFF_DATE = "2023-06-01"

    experiment_query = AttributeQuery(
        attribute="exptl.method",
        operator="in",
        value=EXPERIMENT_TYPES
    )

    query = (
        experiment_query
        & (attrs.entity_poly.rcsb_entity_polymer_type == "Protein")
        & (attrs.entity_poly.rcsb_sample_sequence_length <= length_upper)
        & (attrs.rcsb_accession_info.initial_release_date >= CUTOFF_DATE)
        & (attrs.rcsb_entry_info.resolution_combined <= resolution)
        & (
            (attrs.refine.ls_R_factor_R_free <= r_free)
            | ~attrs.refine.ls_R_factor_R_free.exists()
        )
    )
\end{lstlisting}

After the initial set was fetched from RCSB, additional filtering was conducted using python package gemmi \citep{wojdyr2022gemmi}. Entries  with incomplete residue modeling or missing heavy atoms were flagged, as these indicate gaps in experimental data that could compromise structural accuracy. For peptides with identical sequences across multiple entries, we computed pairwise C$\alpha$-only RMSD using Kabsch alignment; we retained only the best representative and removed conformational outliers with RMSD spread exceeding 3.0\AA. We removed entries containing non-standard residues or histone tags ($\ge$ 5 consecutive histidines), as these indicate modified or artificially-tagged peptides unsuitable for benchmarking. Entries co-crystallized with nucleic acid polymers were also removed. Finally, we removed redundant entries, retaining only the highest-quality representative per sequence-publication group. In cases where the same peptide sequence was deposited under various environments, we chose the best representative based on crystallographic metrics (resolution, R-free) and structural completeness by penalizing peptides with incomplete residues and elevated B-factors relative to the protein background.

\section{PRETZEL-SET: Advanced Dataset Analysis}
\label{sec:set-analysis}

Two decisions in Section~\ref{sec:set} were taken deliberately and both need describing
before the benchmark can be read: we did not exclude peptides chemically close to
molecules already in the pre-cutoff PDB, and we imposed no requirement on the binding
partner. The first makes a calibrated novelty score necessary, because an uncalibrated
similarity between two peptides carries no information. The second makes it necessary to
say what the resulting set actually contains. This appendix does both, in that order.

\subsection{The size-matched null set}
\label{sec:novelty-null}

\paragraph{What the score measures.}
Each peptide chain is compared against every molecule deposited in an entry released before the training cutoff --- 404{,}612 reference molecules, which is the set a co-folding model could have been trained on --- using count-based ECFP4 fingerprints \citep{rogers2010ecfp} and MinMax Tanimoto similarity \citep{swamidass2005minmax}. The
novelty score is $1 - s_{\max}$, the complement of the highest similarity any reference
molecule achieves, so a higher score means further from anything the models saw. Thirty-two
of the 285 chains have a fingerprint-identical match in the pre-cutoff PDB and score zero.

\paragraph{A raw similarity is not interpretable on its own.}
ECFP4 similarity has a high floor between any two peptides: generic backbone amide
environments dominate a peptide's fingerprint, so two unrelated macrocycles of similar size
already agree substantially before any shared chemistry is considered. A score of $0.6$ is
therefore not high or low in itself; it is only meaningful against what a comparable
molecule scores. The null set is what supplies that comparison. It is drawn from real
molecules deposited after the same cutoff and scored against the same pre-cutoff
reference set, so the only thing that differs between the benchmark and its null is which
molecules are being scored.

\paragraph{Why the null is size-matched.}
MinMax Tanimoto is bounded above by $\min(t_q, t_r) / \max(t_q, t_r)$, where $t$ is a
fingerprint's count total, so a similarity is interpretable only between molecules of
comparable size. The benchmark's peptides have a median of 120 heavy atoms (interquartile
range 89--145), with $7.7\%$ at or above 245. The eligible post-cutoff pool has a median of
208 and puts $42\%$ at or above 245.

Size also moves the score. On this reference set larger molecules find closer matches:
within the null, median novelty falls from $0.22$ in the smallest size tertile to $0.08$ in
the largest, most likely because a large post-cutoff chain is usually a protein or nucleic
acid whose fingerprint is dominated by generic backbone environments, and the pre-cutoff
PDB holds hundreds of thousands of those to match it against. A uniform draw from the pool
gives a null with median novelty $0.09$, against $0.15$ for the size-matched one, which
widens the benchmark's margin over the null from $0.16$ to $0.22$. Size matching costs the
benchmark $0.06$ of that margin, and every number below is the matched one.

\paragraph{How the null is drawn, and why n = 1600.}
We cut the benchmark's heavy-atom distribution at its own deciles into ten bins and fill
each bin from the post-cutoff pool in proportion to how many benchmark peptides fall in
it. The null's size histogram then matches the benchmark's bin for bin, every molecule is
drawn once and counted once, and no bin is padded from a neighbour. The size of the sample
is capped by its scarcest bin: on the eligible pool of 14{,}370 molecules that is the
121--127 heavy-atom decile, which the benchmark fills at $8.6\%$ and the pool supplies only
226 molecules for. At $n = 1600$ that bin needs 137, so every decile fills exactly, and the
ceiling before some bin runs short is around 2{,}600. Within that ceiling 1{,}600 is chosen
to be comfortably large: it is five to six null molecules per benchmark peptide, enough to
put the reference curve where it belongs without leaning on the one bin that would break
first.

\paragraph{What the null deliberately does not filter.}
Release date and size are the only two conditions a molecule has to meet to enter the null.
Every other property is left alone, and the result is a mixture of whatever that deposition
window happened to contain: 922 L-peptide chains, 314 DNA and 211 RNA chains, 143 free
ligands, 8 D-peptide chains and 2 nucleic-acid hybrids. Two filters we could have applied are worth naming, because leaving them out was a choice.
We did not remove macrocycles, although 192 of the 1{,}600 ($12\%$) are cyclic. We also did
not remove molecules that had already been deposited before the cutoff and were deposited
again after it; 2{,}790 of the 14{,}370 eligible molecules are such re-depositions. Either
filter would make the null cleaner to describe, and both would break the comparison for the
same reason: the benchmark is not filtered that way either. It is entirely macrocyclic, and
32 of its chains have an exact fingerprint match in the pre-cutoff PDB and were kept.
Screen the null for a property the benchmark keeps, and the gap between the two curves
becomes partly a product of our screening rather than of the chemistry.

\paragraph{Where the fingerprints live.}
The ECFP4 fingerprint matrix for the pre-cutoff PDB is roughly 490\,MB, over GitHub's
per-file limit, so it is not in the code repository; it is deposited on Hugging Face with the
rest of the benchmark's large inputs. The search results it produces are in the
repository, so every figure in this section redraws without it.

\paragraph{The result.}
Against this null the benchmark is appreciably further from the models' training chemistry
than an arbitrary post-cutoff structure of the same size: median novelty $0.31$
(interquartile range $0.19$--$0.39$) against the null's $0.15$. The three cyclization
classes are not equally novel --- head-to-tail macrocycles median $0.35$, stapled $0.35$,
purely disulfide $0.21$ --- which Figure~\ref{fig:characterization}a shows as three
separated cumulative curves.

\subsection{What the benchmark contains}
\label{sec:set-composition}

\begin{figure}[t]
\begin{center}
\includegraphics[width=\linewidth]{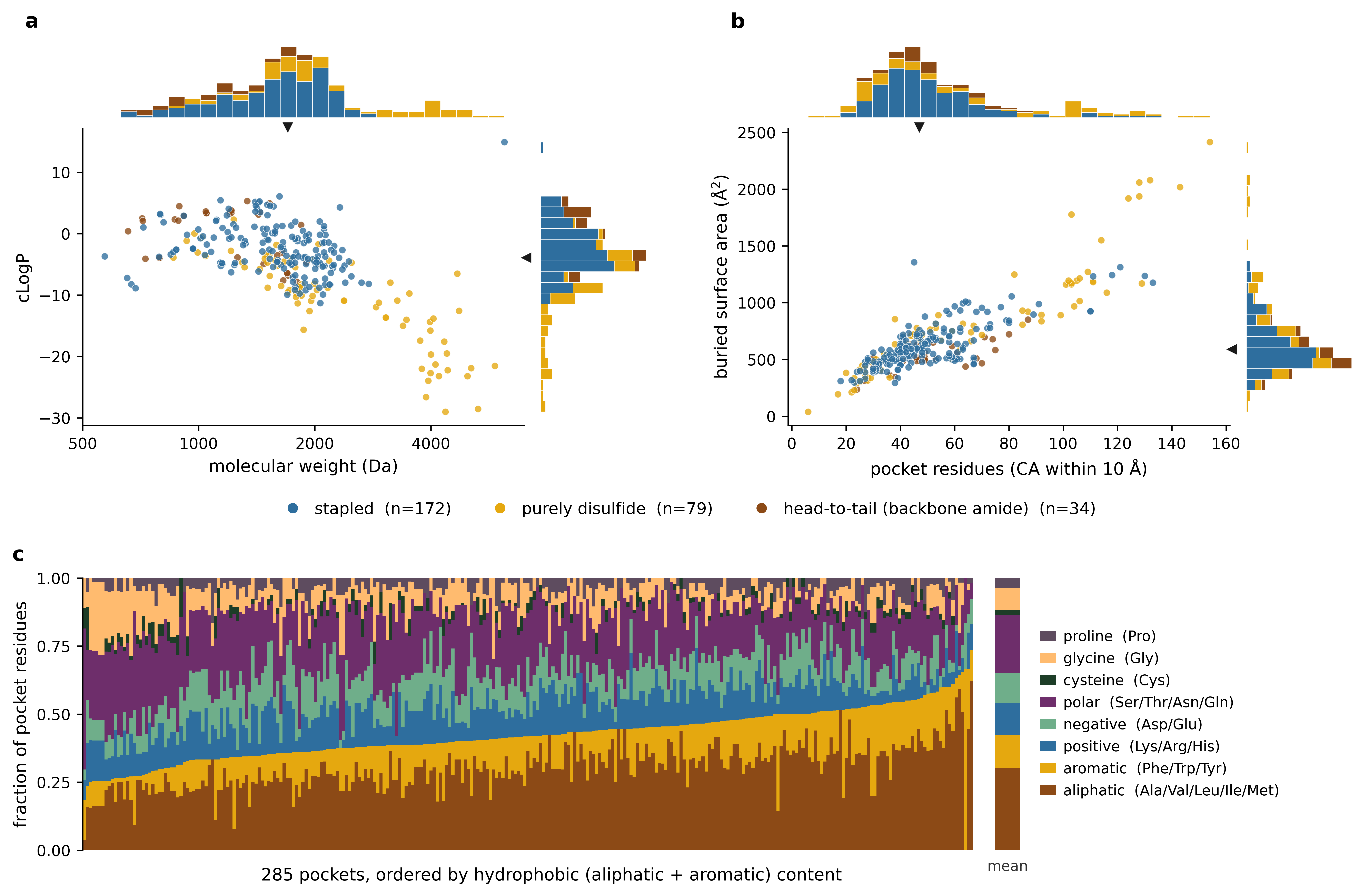}
\end{center}
\caption{Chemical and structural spread of the 285 \textsc{Pretzel-Set} chains and the 285
receptor pockets they bind, coloured throughout by cyclization chemistry.
\textbf{(a)} Peptide molecular weight against cLogP, with stacked marginal histograms
giving each quantity's distribution and its class composition; the triangle on each margin
marks the median of the whole set. \textbf{(b)} The same encoding for the binding site:
number of pocket residues (C$\alpha$ within 10\,\AA{} of the peptide) against buried surface
area. \textbf{(c)} Pocket composition, one stacked column per pocket, ordered left to right
by hydrophobic (aliphatic $+$ aromatic) content. The eight classes partition a pocket, so
every column is exactly full height; the single column at the right is the mean
composition over all 285.}
\label{fig:set-diversity}
\end{figure}

\paragraph{The peptides.}
The set spans 571--6{,}210\,Da, with a median of 1{,}704\,Da over 4--50 residues (median
15). Most of it is chemically modified: 206 of the 285 chains carry at least one
non-canonical residue. cLogP has a median of $-3.92$ and runs from $-29$ to $+15$. The three cyclization classes occupy different regions of
that plane. Head-to-tail macrocycles are the smallest and the only lipophilic group
(median 1{,}215\,Da, cLogP $+2.2$), purely-disulfide peptides are the largest and the most
polar (1{,}992\,Da, cLogP $-9.3$), and stapled peptides sit between them (1{,}659\,Da,
cLogP $-2.7$). The chemistry imposes this: a macrolactam closed head to tail has to be
short enough to close, and a peptide held by disulfides alone has to carry enough cysteines
to do it. The same three colours in Figure~\ref{fig:set-diversity}b are intermixed, so the
chemistry of the peptide says little about the kind of site it binds.

\paragraph{The binding sites.}
Following the pocket definition PXMeter uses for pocket-aligned RMSD
\citep{chen2025pxmeter}, so that characterization and scoring agree, a pocket is the set of
dominant-receptor-chain residues with a C$\alpha$ within 10\,\AA{} of the peptide. Pockets
run from 6 to 154 residues (median 47) and bury 39--2{,}414\,\AA$^2$ of receptor surface
(median 590). The two scale together, as they must. Peptide size predicts buried area much
more weakly: a larger macrocycle does not bury proportionally more receptor, because much
of the extra mass sits away from the interface.

\paragraph{Comparing one pocket with another.}
The map in Figure~\ref{fig:characterization}b needs a number for how alike two binding
sites are, and a pocket cannot supply one on its own: it is a set of residues scattered
through the sequence, a median of 6 separate stretches per site, so two pockets have no
common frame to be laid against each other in. The receptor chains do. Every pair of
complete receptor chains is aligned structurally with foldseek
\citep{vankempen2024foldseek}, which matches them on shape and therefore places residues
opposite one another however far apart they sit in the sequence. Reading that alignment
column by column then answers the question: a column that holds a pocket residue on both
sides is a position the two sites have in common. Pocket coverage is the number of such
columns divided by the number of residues in the first pocket, which is PLINDER's
\texttt{pocket\_qcov} \citep{durairaj2024plinder}.

Dividing by the first pocket's size makes the measure directional --- a small site
contained inside a larger one covers it completely read one way and partly read the other
--- so the map takes the larger of the two values. That matches how PLINDER uses the
metric, where the question is whether a similar site exists anywhere and the best hit is
the answer. The embedding is then over $1 - \texttt{pocket\_qcov}$ as a distance.

\paragraph{What the pockets are made of.}
The eight residue classes partition each pocket: one pocket's eight values sum to $1$, and
they cannot move independently. Figure~\ref{fig:set-diversity}c therefore draws each pocket
as one full-height stacked column, which keeps that constraint visible. Averaged over the
set the composition is $0.30$ aliphatic, $0.21$ polar, $0.12$ aromatic, $0.12$ positive,
$0.11$ negative, $0.08$ glycine, $0.04$ proline and $0.02$ cysteine. Ordering the columns
by hydrophobic content shows the spread: the aliphatic plus aromatic fraction runs from
$0.19$ to $0.74$, so the set holds genuinely hydrophobic and genuinely polar interfaces.
What gives way as hydrophobicity rises is mostly polar and glycine content, leaving the
charged fraction roughly where it was, so the axis of variation across these sites is
polar against greasy.

\section{PRETZEL-TWIST: Definition of cyclization type}
\label{app:definition}

\subsection{Shared conventions}

\paragraph{Cyclization is decided per chain, from the structure.}
Every peptide chain is assigned one of three classes: head-to-tail (backbone
amide), purely disulfide, or stapled. The assignment is derived bond by
bond from the deposited structure rather than from the entry's text annotation.
Three parts of a deposition carry the evidence. The connectivity table
(\texttt{struct\_conn}) lists the links the depositor chose to declare, each one
labelled covalent or disulfide. The polymer table states the chain's residue
order as a polymer, numbered without gaps, which is separate from the residue
numbers printed on the coordinates and often disagrees with them. The Chemical
Component Dictionary (CCD), the reference dictionary of every component code
used in the PDB, supplies each component's atoms, their elements, and whether
the component is an amino acid. Everything below is a rule for reading those
three against each other.

\paragraph{Cross checked.}
The cyclization method is derived twice: once per entry from the declared bonds
alone (initial filter), and once per chain from the structure, the second being what the analysis
uses. The two are written independently and compared on every pipeline run, on PRETZEL-SET they agree on all but three chains, and those three are the case
described in ``not every crosslink is declared'' below.

\subsection{From bonds to cyclization}

\paragraph{Cyclization is a property of the bond graph.}
The chain is first reduced to a graph: residues as nodes, inter-residue bonds as
edges, drawn from three sources --- the bonds the polymer declaration implies,
the bonds the deposition declares, and bonds perceived from interatomic distance
with element and valence checks behind them. The implied polymer bonds have to be
present or no macrocycle is a ring at all, since a closure forms a cycle only
together with the backbone it closes. Every edge is then tested for ring
membership, and those that pass are named by the tests in
Table~\ref{tab:bond-subtypes}, applied in the order shown.

\paragraph{The ring test separates a bridge from a branch.}
A bond lies on a ring when its two residues remain connected to each other after
it is removed. The test is needed because roughly one in fifteen
of the side-chain bonds in PRETZEL-SET hangs off the chain rather than bridging it, and counting those would classify a peptide carrying a single dangling adduct as
stapled.

\paragraph{What a bond is and what a model can do with it are separate questions.}
A linkage is named from chemistry and topology alone. A macrocycle that
closes onto a component the dictionary types as a small molecule is closed head
to tail all the same. Whether a model can build a given bond without being told is recorded separately, per bond, and is used by the config builders, never by the classification.

\begin{table}[h]
\centering
\caption{Linkage tests, applied in this order. ``Adjacent'' means adjacent in the
polymer's own residue order. ``Free at both ends'' means no other peptide bond
reaches that nitrogen and none leaves that carbon.}
\label{tab:bond-subtypes}

\vspace{6pt}
\begin{tabular}{@{}l p{0.52\linewidth} l@{}}
Linkage & Test & Votes for \\[2pt]
\midrule
head-to-tail disulfide & both atoms sulfur; the two residues are the free termini & purely disulfide \\
disulfide              & both atoms sulfur, otherwise                             & purely disulfide \\
backbone               & C and N, adjacent residues                               & none \\
head-to-tail           & C and N, non-adjacent, free at both ends                 & head-to-tail \\
macrocyclic amide      & C and N, non-adjacent, not free at both ends             & stapled \\
terminal cap           & C and N, not on a ring, to a dead-end residue            & none \\
side chain             & any other bond involving an amino acid                   & stapled \\
linker                 & any other bond between two non-amino-acid components     & stapled \\
\end{tabular}
\end{table}
\subsection{Handling edge cases.}

None of the following is an error in the deposition. Each is a place where the
deposited record, read literally, gives the wrong chemistry, and each forced a
decision that the preceding rules encode.

\paragraph{A declared bond is usually not an unusual bond.}
Depositions are required to declare every link that involves a non-standard
component, and a non-canonical macrocycle is mostly non-standard components, so
its ordinary backbone arrives as a list of explicit bond records --- about two
thirds of all declared records in this set. A rule that reads ``declared'' as
``crosslink'' therefore calls almost every peptide here stapled. Backbone bonds
are identified instead by the polymer's own residue order: a carbon-to-nitrogen
bond between residues adjacent in that order is the backbone, whatever the record
says, and it lies on the macrocycle without being what closes it.

\paragraph{Only one of the two residue numberings is chemistry.}
The numbering printed on the coordinates is the depositor's, and it may skip
numbers inside an unbroken chain, distinguish two residues only by a letter
suffix, or run through ligands and cryoprotectant molecules carried in the same
chain. Each of those misleads a different rule: a skipped number breaks the
backbone and makes a cyclic peptide look open, a letter suffix makes two residues
collide into one and loses the staples attached to them, and a glycerol numbered
above the last amino acid becomes the chain's C-terminus. Residue order,
adjacency and the chain's ends are therefore taken from the polymer table, and
every residue is keyed on its number together with its insertion code.

\paragraph{The bond-type label is the depositor's word, not a measurement.}
A record labelled as a generic covalent link is sometimes a genuine
sulfur--sulfur bond, and the atom names a non-canonical residue gives its sulfur
vary (e.g. \texttt{S2} and \texttt{SG}). Chemistry is therefore decided from the elements the CCD assigns, with atom names used only as a fallback for a component missing from the dictionary.

\paragraph{Which amide closes the ring.}
The obvious test --- does an amide join the first and last position of the
sequence? --- fails in both directions, because something other than an amino
acid routinely occupies a sequence end. An acetyl cap takes position one, so the
first position is not the first residue; a pendant residue or a spliced-in linker
takes the last, so a genuine closure does not reach it. The test used instead is
local to the bond: an amide closes the chain head-to-tail when no other peptide
bond reaches that nitrogen and none leaves that carbon, which identifies the
chain's actual ends without assuming where the deposition put them.

\paragraph{Not every crosslink is declared.}
Because bonds are also perceived from coordinates, the per-chain classification
sees crosslinks that the connectivity table omits: five chains of the final set
carry a ring-closing bond that was never declared, each a side-chain crosslink at
bond length for its element pair, and three of them change class as a result.

\paragraph{A bond into a partner chain does not cyclize this one.}
A few peptides carry no intra-chain ring-closing bond and are held to a second
chain by a disulfide. Whether a two-chain molecule stapled to its partner is a
cyclic peptide is a curation question rather than a chemical one; the
classification does not traverse into another chain, so those chains are kept out of the benchmark.

\subsection{Collapsing to three classes}

\paragraph{The three classes follow how the molecules are made.}
The collapse exists to give the shape of the set in one simplified figure, and the cut is drawn along synthetic commitment. A disulfide is the least demanding closure: two cysteines and an oxidation, reversible under reducing conditions, and the backbone is left untouched. A head-to-tail amide is a commitment of a different
order, because it consumes both termini --- changing the peptide's net charge,
removing two hydrogen-bonding groups and with them the exopeptidase liability ---
so the cyclised peptide is a different molecule rather than a decorated one.
Everything else is the chemistry of lead optimisation, where a series is bridged,
re-bridged and linked in whatever way improves it, and the individual tactics are
too numerous and each too sparsely represented to be categories of their own.
One tactic within the stapled class is common enough to be worth mentioning: 93
out of the 172 chains close through a thioether. These peptides are kept inside
the stapled class, but the per-bond typing can be identified using a manifest
explained in Section \ref{sec:engine-examples}.

\paragraph{The same three classes appear on a second, unrelated axis.}
Head-to-tail and disulfide are the only closures any co-folding interface treats
as a case of its own, although how much has to be spelled out varies. The other closures can only be offered as an anonymous pair of atoms, to be honoured, silently removed, or accepted and not built, depending on the model; the next section takes that up. An axis of synthetic effort and an axis of what the tools have a word for cut the set the same way, which is why the summary has three classes and not a finer or a coarser
split.

\paragraph{The collapse.}
Each ring-closing bond of a chain votes for the class in the last column of
Table~\ref{tab:bond-subtypes}; linkages that describe the backbone or a cap vote
for nothing, and are listed as voting for nothing explicitly rather than by
omission. Any vote for stapled wins: a peptide that is closed head to tail
and bridged has been through the lead optimisation step, so it belongs with the optimised ones, and fourteen chains are in that position. Otherwise a chain
voting only head-to-tail, or only disulfide, takes that class --- an interior
disulfide and one between the two termini are different bonds and are recorded as
such, but both leave a peptide cyclised by a disulfide and nothing else. A chain
whose votes are a disulfide and a head-to-tail amide with no bridge, and a chain
with no ring-closing bond at all, are flagged rather than assigned: no policy was
agreed for the mixture, and the absence of a ring-closing bond is a statement
about the chain rather than a classifier failure. Neither case occurs in the
final set, because requiring each chain to close its own ring is also what admits
it to the benchmark.

\paragraph{The third class is a residual, and the largest.}
Defining a class by exclusion puts 172 of the 285 chains in it, so the largest
bar in any figure drawn from this scheme is the one with the loosest definition. It is not formless: 106 of those chains close through a separate linker component, 34 bridge one side chain directly to another, 30 run from a backbone terminus to a side chain, and about three in five of all the bridging bonds involved are carbon--sulfur, so the class is dominated by cysteine alkylation onto a bifunctional linker. Five chains close through an ester onto a side-chain hydroxyl and are counted here as well. These are macrolactones, natural-product-style depsipeptides, and chemically they're the ester analogue of head-to-tail rather than a lead-optimization bridge. Anyone who wants a finer
division has the per-bond linkage of every chain in the released manifest; the three classes are a summary, not the underlying annotation.

\section{PRETZEL-TWIST: Engine Limitations}
\label{sec:limitation}
For HighFold3, we observed that explicitly specified disulfide constraints were not reliably encoded for peptides containing non-standard residues, due to a mismatch between the two indexing conventions used internally by the implementation. The disulfide bond is set using one-based residue positions within a chain. Internally, however, the cyclic position offset encoding matrix (CycPOEM) is constructed over the token axis inherited from the AlphaFold 3 featurisation pipeline. In that pipeline, canonical amino acids are represented by a single representative-atom token (typically C$\alpha$), while non-standard residues and ligand entities are atomized into one token per heavy atom. The chain length used to allocate the CycPOEM is derived from the token array, while the user-supplied cysteine positions are applied directly as row and column indices into that matrix without an intervening residue-to-token mapping. Consequently, for a chain containing a non-standard residue with N heavy atoms at position k, all residues downstream of k are displaced by N - 1 positions in the token index space, and the specified bond is written to a matrix element that does not correspond to the intended residue pair. The two index spaces coincide only when every residue in the chain is canonical, which is presumably why the behavior is not apparent for standard peptides.

Given these considerations, we elected to remove all explicitly specified disulfide bonds from our HighFold3 prediction runs, and to rely on the model's implicit treatment of disulfide geometry. We retained the cyclization matrix for head-to-tail bond specification across both stapled and head-to-tail cyclization classes.


\section{PRETZEL-TWIST: Engine examples}
\label{sec:engine-examples}

One record per chain in \texttt{peptide\_manifest.json} is the whole description
of a peptide that the config builders read, and each of the four builders turns
that record into its own model's input. The record is written so that no builder
has to re-derive any chemistry: it states what the chain contains and also why
each residue and each bond is treated the way it is. Two chains are shown below
with all four configurations, one chain that needs nothing unusual and one that
needs everything. Long sequences and the inline dictionary block are elided,
the alignment and template fields each format carries are left out because they
say nothing about the chemistry, and YAML sequences are printed in flow style to
save space. Every other field is as generated.

\subsection{9KKU-C, where the chemistry is simple and two formats still cannot
express it}

\paragraph{Manifest record.}\leavevmode

\begin{lstlisting}[style=cfg]
{ "pdb": "9KKU", "chain": "C",
  "category": "single chain",
  "cyclization_methods": ["head to tail (disulfide)"],
  "foldable_as_single_chain": true,
  "n_segments": 1, "n_ligand_entities": 0,
  "chosen_among_copies": ["C", "D"],
  "bond_subtypes":     {"head-to-tail disulfide": 1},
  "bond_requirements": {"protein-protein, same chain": 1},
  "declared_bonds": ["CYS@1.SG-CYS@43.SG [head-to-tail disulfide; found by disulf]"],
  "implicit_backbone_bonds": 42,
  "target_chains": ["A", "B"],
  "build": {
    "segments": [{"positions": [1, ..., 43],
                  "sequence": "CAAELAALEAELAALEG...LQFLIKKLKQLKVAC",
                  "modifications": []}],
    "ligands": [] } }
\end{lstlisting}

\paragraph{Protenix config.}\leavevmode

\begin{lstlisting}[style=cfg]
{ "name": "9KKU_C_seq_Protenix",
  "sequences": [
    {"proteinChain": {"sequence": "CAAELAALEAELAALEG...LKQLKVAC",
                      "count": 1, "id": ["H"]}},
    {"proteinChain": {"sequence": "GQNHHEVVKFMDVYQRSY...KCECRPKKD",
                      "count": 1, "id": ["A"]}},
    {"proteinChain": {"sequence": "GQNHHEVVKFMDVYQRSY...KCECRPKKD",
                      "count": 1, "id": ["B"]}}],
  "covalent_bonds": [
    {"entity1": 1, "position1": 1,  "atom1": "SG",
     "entity2": 1, "position2": 43, "atom2": "SG"}] }
\end{lstlisting}

\paragraph{Boltz-2 config.}\leavevmode

\begin{lstlisting}[style=cfg]
name: 9KKU_C
model: boltz2
model_seed: 1
sequences:
- protein:
    id: P1
    sequence: CAAELAALEAELAALEG...LQFLIKKLKQLKVAC
- protein:
    id: A
    sequence: GQNHHEVVKFMDVYQRSY...KCECRPKKD
- protein:
    id: B
    sequence: GQNHHEVVKFMDVYQRSY...KCECRPKKD
constraints:
- bond:
    atom1: [P1, 1, SG]
    atom2: [P1, 43, SG]
\end{lstlisting}

\paragraph{AF3 config.}\leavevmode

\begin{lstlisting}[style=cfg]
{ "name": "9KKU_C", "dialect": "alphafold3", "version": 1,
  "modelSeeds": [1],
  "sequences": [
    {"protein": {"id": "A", "sequence": "CAAELAALEAELAALEG...LKQLKVAC"}},
    {"protein": {"id": "B", "sequence": "GQNHHEVVKFMDVYQRSY...CECRPKKD"}},
    {"protein": {"id": "C", "sequence": "GQNHHEVVKFMDVYQRSY...CECRPKKD"}}] }
\end{lstlisting}

\paragraph{HighFold3 config, and the run line that goes with it.}\leavevmode

\begin{lstlisting}[style=cfg]
{ ... character for character the AF3 config above ... }

run_alphafold.py --json_path=.../9KKU_C_HighFold3.json \
                 --output_dir=.../9KKU_C --head_to_tail=0
\end{lstlisting}

\paragraph{The one cyclizing bond reaches two models out of four.}
Boltz-2 takes it as a bond constraint and Protenix as a \texttt{covalent\_bonds}
record. The AF3 and HighFold3 configs contain no bond at all, for two different
reasons. AF3's format does not accept a bond between two polymer residues and
removes such bonds before inference, so the builder does not write it and logs it
instead as one unsupported polymer-polymer bond. HighFold3 does accept them, but
through command-line flags rather than through the config, and the only flag that
fits this bond is the one for intra-chain disulfides, which this pipeline stopped
using because it mis-indexes whenever a non-canonical residue lies between the two
cysteines. Two of the four models are therefore shown a linear peptide.

\paragraph{Both copies of the receptor are declared.}
The record names two target chains for this peptide because it touches both, so
both are emitted, and a prediction is scored against the pair rather than against
a single copy chosen for convenience. The next chain is the other side of the
same rule.

\subsection{10SB-D, a chain that needs everything}

\paragraph{Manifest record.}\leavevmode

\begin{lstlisting}[style=cfg]
{ "pdb": "10SB", "chain": "D",
  "category": "multi chain",
  "cyclization_methods": ["stapled"],
  "foldable_as_single_chain": false,
  "n_segments": 2, "n_ligand_entities": 5,
  "residues": {
    "SIN@1":   "type NON-POLYMER; terminal, treated as ligand",
    "2RA@3":   "inline CCD modification",
    "A1CHA@4": "type non-polymer; interior, treated as ligand and chain split",
    "FTR@5":   "inline CCD modification",
    "0A1@8":   "inline CCD modification",
    "3WX@9":   "type peptide-like; terminal, treated as ligand",
    "A1C8P":   "non-polymer fragment; bonded ligand entity",
    "GOA":     "non-polymer fragment; bonded ligand entity"},
  "bond_subtypes":     {"side-chain": 3, "linker": 3,
                        "macrocycle backbone": 3},
  "bond_requirements": {"protein-ligand": 6, "ligand-ligand": 3},
  "declared_bonds": [
    "SIN@1.C1-ALA@2.N     [side-chain;          protein-ligand]",
    "SIN@1.C4-A1C8P.N4    [linker;              ligand-ligand ]",
    "2RA@3.C-A1CHA@4.N    [macrocycle backbone; protein-ligand]",
    "A1CHA@4.C-FTR@5.N    [macrocycle backbone; protein-ligand]",
    "A1CHA@4.N1-GOA.C     [linker;              ligand-ligand ]",
    "FTR@5.NE1-A1C8P.C19  [side-chain;          protein-ligand]",
    "PRO@6.CB-GOA.O2      [side-chain;          protein-ligand]",
    "0A1@8.C-3WX@9.N      [macrocycle backbone; protein-ligand]",
    "3WX@9.C-A1C8P.N5     [linker;              ligand-ligand ]"],
  "implicit_backbone_bonds": 4,
  "target_chains": ["B"],
  "non_contacting_chains_not_declared": ["A"],
  "build": {
    "segments": [{"positions": [2, 3],       "sequence": "AX",
                  "modifications": [[2, "2RA"]]},
                 {"positions": [5, 6, 7, 8], "sequence": "WPTY",
                  "modifications": [[1, "FTR"], [4, "0A1"]]}],
    "ligands":  [{"comp_id": "SIN",   "position": 1},
                 {"comp_id": "A1CHA", "position": 4},
                 {"comp_id": "3WX",   "position": 9},
                 {"comp_id": "A1C8P", "position": null},
                 {"comp_id": "GOA",   "position": null}] } }
\end{lstlisting}

\paragraph{Protenix config.}\leavevmode

\begin{lstlisting}[style=cfg]
{ "name": "10SB_D_seq_Protenix",
  "sequences": [
    {"proteinChain": {"sequence": "AX", "count": 1, "id": ["H"],
                      "modifications": [{"ptmType": "CCD_2RA",
                                         "ptmPosition": 2}]}},
    {"proteinChain": {"sequence": "WPTY", "count": 1, "id": ["I"],
                      "modifications": [{"ptmType": "CCD_FTR",
                                         "ptmPosition": 1},
                                        {"ptmType": "CCD_0A1",
                                         "ptmPosition": 4}]}},
    {"proteinChain": {"sequence": "SIPWNLERITPPRYRA...GAGNSHHHHHH",
                      "count": 1, "id": ["A"]}},
    {"ligand": {"ligand": "CCD_SIN",   "count": 1, "id": ["B"]}},
    {"ligand": {"ligand": "CCD_A1CHA", "count": 1, "id": ["C"]}},
    {"ligand": {"ligand": "CCD_3WX",   "count": 1, "id": ["D"]}},
    {"ligand": {"ligand": "CCD_A1C8P", "count": 1, "id": ["E"]}},
    {"ligand": {"ligand": "CCD_GOA",   "count": 1, "id": ["F"]}}],
  "covalent_bonds": [
    {"entity1": 4, "position1": 1, "atom1": "C1",
     "entity2": 1, "position2": 1, "atom2": "N"},
    {"entity1": 1, "position1": 2, "atom1": "C",
     "entity2": 5, "position2": 1, "atom2": "N"},
    {"entity1": 5, "position1": 1, "atom1": "C",
     "entity2": 2, "position2": 1, "atom2": "N"},
    ... six more, one per declared bond ...] }
\end{lstlisting}

\paragraph{Boltz-2 config.}\leavevmode

\begin{lstlisting}[style=cfg]
name: 10SB_D
model: boltz2
model_seed: 1
sequences:
- protein:
    id: P1
    sequence: AX
    modifications: [{position: 2, ccd: 2RA}]
- protein:
    id: P2
    sequence: WPTY
    modifications: [{position: 1, ccd: FTR}, {position: 4, ccd: 0A1}]
- protein:
    id: B
    sequence: SIPWNLERITPPRYRA...GAGNSHHHHHH
- ligand: {id: L1, ccd: SIN}
- ligand: {id: L2, ccd: A1CHA}
- ligand: {id: L3, ccd: 3WX}
- ligand: {id: L4, ccd: A1C8P}
- ligand: {id: L5, ccd: GOA}
constraints:
- bond: {atom1: [L1, 1, C1], atom2: [P1, 1, N]}
- bond: {atom1: [P1, 2, C],  atom2: [L2, 1, N]}
- bond: {atom1: [L2, 1, C],  atom2: [P2, 1, N]}
  ... six more, one per declared bond ...
\end{lstlisting}

\paragraph{AF3 config.}\leavevmode

\begin{lstlisting}[style=cfg]
{ "name": "10SB_D", "dialect": "alphafold3", "version": 1,
  "modelSeeds": [1],
  "sequences": [
    {"protein": {"id": "A", "sequence": "AX",
                 "modifications": [{"ptmType": "2RA", "ptmPosition": 2}]}},
    {"protein": {"id": "B", "sequence": "WPTY",
                 "modifications": [{"ptmType": "FTR", "ptmPosition": 1},
                                   {"ptmType": "0A1", "ptmPosition": 4}]}},
    {"protein": {"id": "C", "sequence": "SIPWNLERITPPRYRA...GAGNSHHHHHH"}},
    {"ligand":  {"id": "D", "ccdCodes": ["SIN"]}},
    {"ligand":  {"id": "E", "ccdCodes": ["A1CHA"]}},
    {"ligand":  {"id": "F", "ccdCodes": ["3WX"]}},
    {"ligand":  {"id": "G", "ccdCodes": ["A1C8P"]}},
    {"ligand":  {"id": "H", "ccdCodes": ["GOA"]}}],
  "bondedAtomPairs": [
    [["D", 1, "C1"], ["A", 1, "N"]],
    [["A", 2, "C"],  ["E", 1, "N"]],
    [["E", 1, "C"],  ["B", 1, "N"]],
    ... six more, one per declared bond ...] }
\end{lstlisting}

\paragraph{HighFold3 config, and the run line that goes with it.}\leavevmode

\begin{lstlisting}[style=cfg]
{ ... the AF3 config above, entity for entity and bond for bond, plus: ...
  "userCCD": "data_SIN ... 59,730 characters of component definitions ..." }

run_alphafold.py --json_path=.../10SB_D_HighFold3.json \
                 --output_dir=.../10SB_D --head_to_tail=0
\end{lstlisting}

\paragraph{The peptide is two protein entities because one residue cannot be a
residue.}
\texttt{A1CHA} at position 4 is typed in the Chemical Component Dictionary as a
small molecule, and no input format allows a small molecule to sit inside a
polymer chain. It becomes a ligand entity of its own and the peptide is cut at
it, leaving positions 2--3 and 5--8 as two protein entities. The manifest states
that conclusion in its \texttt{residues} map next to the reason for it, so every
builder makes the same cut and the audit can check that it did.

\paragraph{A bond can be ordinary chemistry and still have to be declared.}
Three of the nine bonds are ordinary peptide bonds, \texttt{2RA@3.C-A1CHA@4.N}
and the two like it. The linkage naming used elsewhere in this appendix calls
them \texttt{macrocycle backbone}, and they vote for no cyclization class,
because chemically they are just backbone. They are nonetheless written into
every config, because the cut has put their two ends in different entities and no
model builds a bond across an entity boundary for free. What a bond is and what
the model has to be told about it are different questions, and the manifest
answers both for every bond rather than only one of them.

\paragraph{Positions are renumbered, and the manifest carries the mapping.}
\texttt{2RA} is residue 3 of the deposition and position 2 of the first entity;
\texttt{FTR} is residue 5 and position 1 of the second; \texttt{0A1} is residue 8
and position 4. Both numberings are in the record, \texttt{positions} in the
deposition's numbering and \texttt{modifications} indexed within the segment,
which is what lets a builder emit a config and lets the fidelity check read that
config back to the deposition without guessing.

\paragraph{Here all four formats accept every bond.}
The nine bonds appear as \texttt{bondedAtomPairs} in AF3 and HighFold3, as
\texttt{constraints} in Boltz-2 and as \texttt{covalent\_bonds} in Protenix, and
none is dropped. This is the opposite of the previous chain, and the difference
is not complexity: 10SB-D is the harder molecule by every other measure, but not
one of its bonds joins two polymer residues, which is the single case the formats
disagree about. HighFold3 additionally carries the definitions of the
non-standard components inline in a \texttt{userCCD} block, so a run does not
depend on which dictionary version is installed beside the model.

\paragraph{Only contacting receptor chains are declared.}
This entry contains two copies of the protease and only one of them touches the
peptide. The other is named in the record as excluded for lack of contact. The contacting copy is chain B of the deposition and is seated as chain A of the config. The record keeps the deposited chain identifier, which is what lets a prediction be read back against the structure it came
from.

\section{PRETZEL-SALT: Additional Accuracy Metrics}
\label{app:accuracy}
\emph{Interface RMSD (iRMSD)} is the CAPRI criterion \citep{janin2003capri}: the residues forming the
interface in the reference are superposed and scored using the same atoms, so
the rigid-body component of the error is removed by construction. \emph{Backbone RMSD}
keeps that component --- it inherits the pocket alignment of the main text and
merely restricts scoring to N, C$\alpha$, C and O. The two therefore suggest the main-text metric in different ways, and the gap between them identifies a prediction with the right binding mode in the wrong place.

\paragraph{Protocol.}
The interface is read off the reference only: a residue qualifies if any heavy
atom lies within 10~\AA{} of any heavy atom on the opposite side, with
membership all-or-nothing per residue. Backbone atoms (N, C$\alpha$, C, O) of
every recruited residue on both sides are scored. Receptor residues are paired
by identity, peptide residues by the same graph-matched correspondence used for
the pocket-aligned RMSD, so all three metrics describe one atom pairing. For
iRMSD a single rigid transform is fit on the union of those atoms, reflections
disallowed; backbone RMSD reuses the pocket transform unchanged. Degenerate
interfaces are reported as undefined rather than 0.0; no target required this.

\begin{table}[htbp]
  \centering
  \caption{Interface and backbone RMSD (\AA) for 285 targets per method. Best
    value in each column in bold. No pairwise difference in the iRMSD columns is
    statistically significant (McNemar on paired targets, all $p \ge 0.19$).}
  \label{tab:irmsd}
  \begin{tabular}{l rrr rrr}
    \hline
    & \multicolumn{3}{c}{Interface RMSD} & \multicolumn{3}{c}{Backbone RMSD} \\
    \cline{2-4}\cline{5-7}
    Method & Median & $<$2~\AA & $<$5~\AA & Median & $<$2~\AA & $<$5~\AA \\
    \hline
    AlphaFold3  & 3.87          & 107 (37.5\%)          & 164 (57.5\%)          & 9.57          & 64 (22.5\%)          & 104 (36.5\%) \\
    Boltz-2     & \textbf{3.27} & 110 (38.6\%)          & 165 (57.9\%)          & 8.55          & 55 (19.3\%)          & 109 (38.2\%) \\
    Protenix-v2 & 3.36          & \textbf{117} (41.1\%) & \textbf{170} (59.6\%) & \textbf{8.39} & \textbf{70} (24.6\%) & \textbf{117} (41.1\%) \\
    HighFold3   & 3.75          & 109 (38.2\%)          & 161 (56.5\%)          & 8.80          & \textbf{70} (24.6\%) & 110 (38.6\%) \\
    \hline
  \end{tabular}
\end{table}

The four methods are very similar on iRMSD. On backbone RMSD
Protenix-v2 leads, and tells a similar story with our original RMSD calculation.

\paragraph{Why these are secondary metrics.}
Both report smaller numbers than the pocket-aligned peptide RMSD, but only one of them changes what is being measured.

Backbone RMSD is a mild relaxation: it scores 56 of a median 120 peptide heavy atoms under the same transform, which lowers the median by a factor of 1.12 and leaves the ranking essentially unchanged ($\rho = 0.992$).

iRMSD is not a relaxation but a different quantity. Its scored atom set is dominated by the receptor --- a median of 248 receptor against 56 peptide backbone atoms, a ratio $R = 5.2$, with $R > 3$ on 81\% of rows --- and the receptor superposes well whatever the peptide does. Writing $r$ for the receptor
residual and $d$ for the peptide displacement, $r \ll d$ gives
\begin{equation}
  \label{eq:irmsd-compression}
  \mathrm{iRMSD} = \sqrt{\frac{N_r r^{2} + N_p d^{2}}{N_r + N_p}}
    \approx \frac{d}{\sqrt{R + 1}},
\end{equation}
a predicted compression of 2.50 against an observed median ratio of 2.63. An
iRMSD of 3.2~\AA{} does not mean the peptide is 3.2~\AA{} from native.

The refit also discards placement error by design, and this is where the two
metrics genuinely disagree. Seventy-nine rows fall below 2~\AA{} iRMSD ---
nominally CAPRI ``medium accuracy'' --- with peptide RMSD above 5~\AA. On 9NLO
chain~C, Boltz-2 scores 11.15~\AA{} peptide RMSD against 1.37--1.76~\AA{} for
the other three, yet its iRMSD (1.47~\AA) is comparable to theirs
(0.64--1.08~\AA); all seven peptide residues are scored, so this is not a
coverage artifact. Rank correlation is high ($\rho = 0.959$), but these
disagreements are the informative cases.

Finally, no coverage guard is applied to the interface pairing. On five targets
(8QYA\_A, 8H8Q\_A, 9Q1L\_F, 8DS7\_M, 9I2B\_U) only 11--44\% of the expected
receptor interface backbone could be paired, and the values are reported without
a warning. All four methods are affected identically and every value lies far
above both thresholds, so the comparison is unaffected.

\section{PRETZEL-SALT: Chemical Validity Metrics}
\label{sec:chem-validity}

\subsection{Shared conventions}

\paragraph{Chemistry is perceived once, in the deposition.}
Bonded atom pairs, amide typing, carbon hybridisation and ring membership are
all decided from the deposited coordinates (plus \texttt{struct\_conn} links).
The prediction is then measured at exactly those atoms and its chemistry is
never re-perceived. Independent perception would hide errors: a bond stretched
to 4\,\AA{} would no longer be perceived as a bond and so would never be
measured, and a flattened tetrahedral carbon would be re-typed as trigonal and
scored against the $120^\circ$ band it now satisfies.

\paragraph{Validity rule.}
A measurement $x$ is valid if it lies inside its literature reference band
\emph{or} reproduces the deposited value:
\begin{equation}
  \text{valid}(x) \;=\;
  \bigl[x \in \text{band}\bigr] \;\lor\;
  \bigl[\,|x_{\text{pred}} - x_{\text{dep}}| \le \tau\,\bigr],
\end{equation}
with $\tau = 0.05$\,\AA{} for bonds, $5^\circ$ for angles, $0.05$\,\AA{} for
out-of-plane distances, and the class tolerance for torsions. The first clause
prevents the metric from collapsing into ``copy the answer'': a prediction with
an ordinary 1.33\,\AA{} amide passes even if the deposition refined that amide
at 1.45\,\AA. The second clause makes the check ground-truth-referenced, so a
prediction is never penalised for agreeing with the structure it was asked to
predict.

\paragraph{Atom correspondence.}
All evaluated models retain CCD component ids and atom names, so atoms are
paired by residue: the receptor is identified by sequence similarity (required
to align as a single contiguous run, and never a sister copy of the peptide)
and set aside, and the remaining residues are paired by component id in order.
The mapping is verified rather than trusted: for every chain with a prediction,
all 4{,}738 deposited inter-residue bonds are located in the prediction, with
no partial matches.

\subsection{Ground-truth control: the deposition passes every test}
\label{sec:gt-control}

As a control, each deposited structure is scored against itself, taking the
deposition as both reference and prediction. The atom mapping is then the
identity, so every measurement is located and none is dropped. The control
passes all four metrics on every chain, and this follows directly from how the
metrics are defined.

\subsection{Metric 1: Bond distance}

\paragraph{Implementation.}
Every inter-residue bond is measured as the interatomic distance between its
two atoms, in both structures (4{,}738 bonds over 285 chains). Bonds are enumerated from three sources: explicit \texttt{struct\_conn} records, polymer
backbone bonds inferred from chain connectivity, and remaining crosslinks --- frequently
unannotated for synthetic chemistries --- perceived from coordinates with a covalent-radius
criterion. Because mmCIF bond-order annotation is unpopulated in nearly all depositions, types
spanning two chemistries with distinct equilibrium lengths are given a single interval
admitting both, which reduces sensitivity for those types without producing false failures. The
interval-only pass rate, omitting the deposited-value clause of Eq.~2, is retained as a
diagnostic on the intervals themselves. Each bond is
assigned one of 21 bond types with a literature band
(Table~\ref{tab:bond-bands}); 92.3\% are C--N amides, so the column is
predominantly a statement about peptide bonds. Because connectivity is
perceived from coordinates, bond orders are unavailable and amide typing is
geometric: a C--N bond is an amide when the carbon carries an O within
1.32\,\AA{} or an S within 1.75\,\AA, or when the carbon is planar and carries
any oxygen. Element pairs with no defined band fall to a catch-all type whose
in-band value is not meaningful; every such bond is listed in the report.

\begin{table}[t]
\centering
\caption{Reference bands for the most frequent inter-residue bond types. A bond passes
when its predicted length falls inside the band or reproduces the deposited value to
within 0.05\,\AA.}
\label{tab:bond-bands}
\vspace{0.5em}
\small
\setlength{\tabcolsep}{3.5pt}
\begin{tabular}{@{}l cc l@{}}
\hline
Bond type & Ideal (\AA) & Band (\AA) & Chemistry / source \\
\hline
  C--N amide            & 1.336 & 1.28--1.40 & Engh \& Huber \\
  C--S                  & 1.800 & 1.65--1.92 & pooled thioether / thioester / thiazole \\
  S--S                  & 2.040 & 1.95--2.13 & CSD consensus \\
  C--C                  & 1.520 & 1.44--1.62 & single bond, $\ge 1$ sp$^3$ carbon \\
  C--N (non-amide)      & 1.470 & 1.40--1.55 & amine / $N$-alkyl single bond \\
  C--O                  & 1.420 & 1.29--1.50 & ether or ester single bond \\
  C(sp$^2$)--C(sp$^2$)  & 1.430 & 1.28--1.58 & pooled alkene / conjugated / biaryl \\
\hline
\end{tabular}
\end{table}
\subsection{Metric 2: Bond angles}

\paragraph{Implementation.}
Every angle at an inter-residue junction is computed as the angle at the
vertex atom between its two neighbours. Bands come from two sources. About 91\% of angles fall on named bands
backed by restraint-library values (Table~\ref{tab:angle-bands}). The remaining
9\% fall on generic bands keyed to the vertex element and local geometry
(e.g.\ C~sp$^3$ 99--120$^\circ$, C~sp$^2$ 110--130$^\circ$, N~sp$^2$
110--130$^\circ$, S~sp$^3$ 90--112$^\circ$), which carry no specific citation.
The vertex geometry class is decided in the deposition, so a predicted centre
is always scored against the band of the hybridisation it should have.

\begin{table}[t]
\centering
\caption{Named restraint bands for inter-residue bond angles. Angles without a named
restraint are scored against the ideal for the vertex atom's hybridization.}
\label{tab:angle-bands}
\vspace{0.5em}
\small
\setlength{\tabcolsep}{3.5pt}
\begin{tabular}{@{}l cc l@{}}
\hline
Angle & Ideal ($^\circ$) & Band ($^\circ$) & Source \\
\hline
  O--C--N    & 122.7 & 115--130 & Engh \& Huber \\
  CA--C--N   & 116.2 & 108--124 & Engh \& Huber \\
  C--N--CA   & 121.7 & 113--130 & Engh \& Huber \\
  CB--SG--SG & 103.8 & 94--114  & CSD consensus \\
\hline
\end{tabular}
\end{table}

\subsection{Metric 3: Dihedral}

\paragraph{Implementation.}
Three classes of measurement enter this column
(Table~\ref{tab:dihedral-classes}). Omega accepts both $180^\circ$ and
$0^\circ$ because a \emph{cis} peptide bond is correct chemistry (the benchmark
contains 186). Amide planarity is the distance of the carbonyl carbon from the
plane of its substituents, expressed in \aa ngstr\"om so that backbone and
macrocyclising amides share one scale. C$\alpha$ chirality is the improper
torsion N--C--CA--CB. Angular classes are compared on the circle
($-179^\circ$ is $1^\circ$ from $180^\circ$); planarity is compared linearly.

\begin{table}[t]
\centering
\caption{Dihedral-class measurements. Each is scored against the nearest of its
permitted targets, which are plural because \emph{cis} amides and D-residues are
correct chemistry.}
\label{tab:dihedral-classes}
\vspace{0.5em}
\small
\setlength{\tabcolsep}{3.5pt}
\begin{tabular}{@{}ll cc@{}}
\hline
Class & Quantity & Target & Tolerance \\
\hline
  $\omega$              & rotation about the peptide bond           & $180^\circ$ or $0^\circ$         & $\pm 30^\circ$ \\
  Carbonyl planarity    & C displacement from its substituents' plane & 0\,\AA                         & 0.10\,\AA \\
  CA improper magnitude & improper N--C--CA--CB                     & $\pm 122^\circ$                  & $\pm 18^\circ$ \\
\hline
\end{tabular}
\end{table}

A fourth class, the CB--SG--SG--CB disulfide torsion ($\pm 90^\circ \pm
40^\circ$), was removed: a disulfide away from $\pm 90^\circ$ is strained but
is a conformational preference rather than a validity criterion. The disulfide
bond length and CB--SG--SG angle remain checked under Metrics~1 and~2.

For chirality, the nearer of $\pm 122^\circ$ is currently taken as the target,
so the sign of the improper relative to the deposition is computed (written as
\texttt{sign\_flipped}) but does not enter the verdict. The intended
refinement is to require sign agreement with the deposition, gated on
$|\text{deposited}| > 90^\circ$ so that $\beta$-amino acids, where the improper
is not a stereo descriptor, are excluded.

\subsection{Metric 4: Ring closure}

\paragraph{Implementation.}
This metric detects a macrocycle whose individual bonds are ideal but which
was built as an open chain. The procedure is:
\begin{enumerate}
  \item Perceive bonds in the deposition from coordinates and
        \texttt{struct\_conn} links.
  \item Find rings as the smallest cycle through each bond: remove the bond,
        run a breadth-first search for the shortest remaining path between its
        two atoms, and take that path plus the bond as the ring. Rings are
        deduplicated by edge set.
  \item Keep macrocycles only, defined structurally as rings spanning more than
        one residue. Rings inside a single component come from the CCD and are
        not predicted.
  \item Look up every ring bond in the prediction. A ring with any unmapped
        atom is unmeasurable and is dropped; it never counts as a failure.
  \item A chain passes when no macrocycle in it contains a broken bond.
\end{enumerate}
A bond of length $d$ is \emph{broken} when
$d > \text{band}_{\text{hi}} + 0.50$\,\AA{} and
$d > d_{\text{dep}} + 0.50$\,\AA; the threshold is relative to bond type
because an absolute cutoff that flags a 1.33\,\AA{} amide would sit barely
past a healthy diselenide. Collapsed bonds ($d < 1.25$\,\AA{} and
$d < d_{\text{dep}} - 0.30$\,\AA) are excluded from this column, since they
are already caught by Metric~1; a stricter variant including them is reported
separately.

\section{Experiments: Baseline study with HighFold3 Targets}
\label{sec:baseline}
In order to confirm the quality of our config builders, we have evaluated 17 complexes collected from the HighFold Dataset (1SFI, 3AV9, 3AVA, 3AVB, 3AVF, 3AVG, 3AVH, 3AVI, 3AVJ, 3AVK, 3AVM, 3AVN, 3P8F, 3WNE, 3ZGC, 4K1E, 4KEL). These were all deposited to the RCSB before the training cutoff date and are, therefore, not a subset of our \textsc{PRETZEL-SET}.

Using \textsc{PRETZEL-TWIST}, we prepared input configurations for HighFold3 for the 17 complexes. 14 of the 17 complexes contain head-to-tail cyclic peptides and the remaining are head-to-tail cyclic and also contain an inner disulfide bond. After prediction, we get a median RMSD of 1.02 \AA\ with 16 out of 17 complexes being below 2 \AA, which exposes the difficulty in our dataset and also alludes to issues in interpreting benchmarking results where the model has trained on the predicted structure. The HighFold3 paper reports a median RMSD of 0.266 \AA; they calculate RMSD by evaluating only C$\alpha$ atoms and superposing the peptide onto itself without considering the receptor. As a check, we implemented their calculation procedure and got 0.208 \AA, further demonstrating that our configs are accurately replicating their benchmark work.

\section{Experiments: Case Study}
\label{sec:case-study}

\subsection{9EMZ: ordinary bonding, non-canonical residues}
\label{sec:case-9emz}

\begin{figure}[t]
\centering
\includegraphics[width=\linewidth]{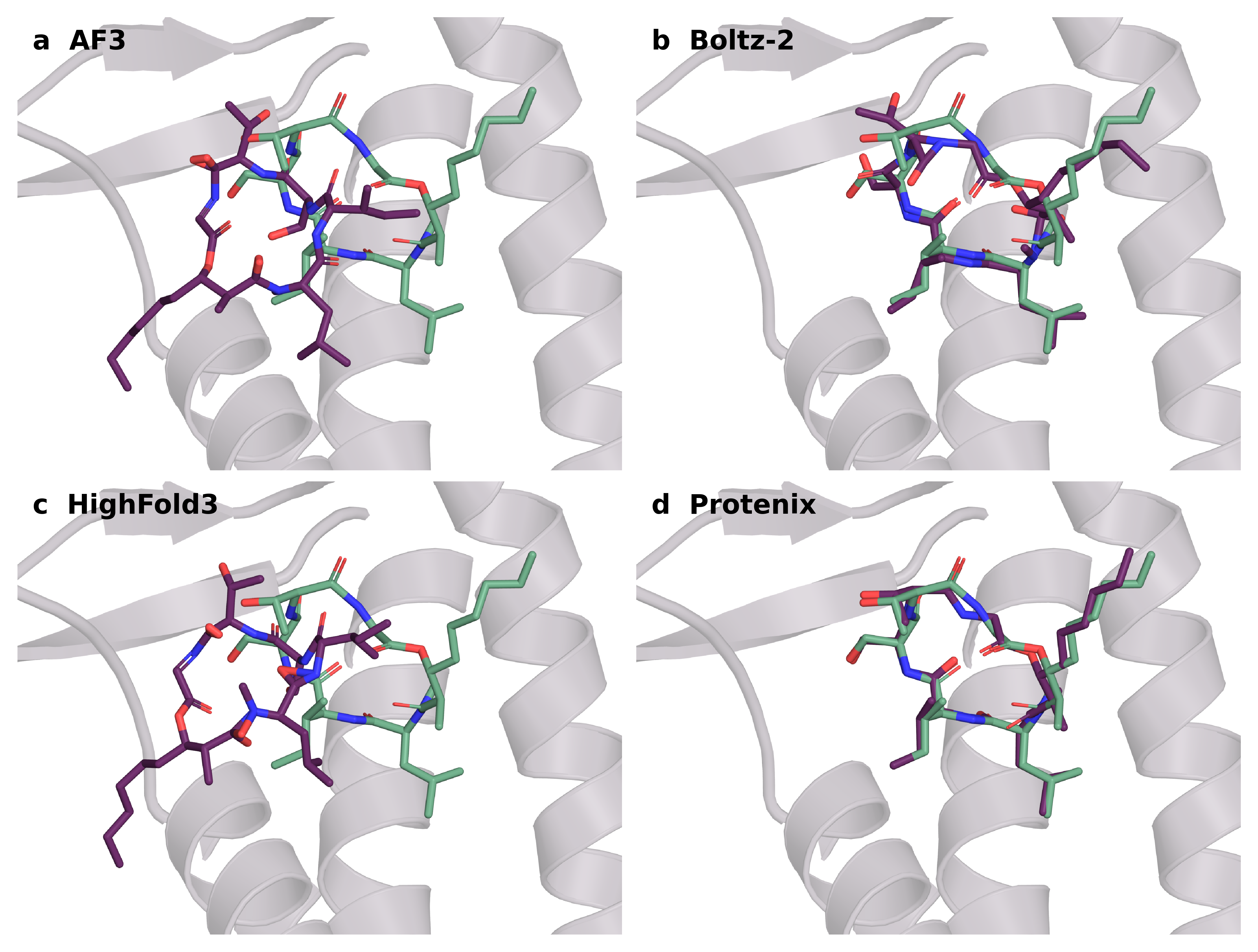}

\vspace{6pt}
{\small
\setlength{\tabcolsep}{3.5pt}
\begin{tabular}{@{}l cccc c cc@{}}
\hline
& \multicolumn{5}{c}{Chemical validity} & \multicolumn{2}{c}{RMSD (\AA)} \\
\cline{2-6}\cline{7-8}
Model & \makecell{Bond\\distance} & \makecell{Bond\\angles} & Dihedral & \makecell{Ring\\closure}
      & \makecell{Checks\\passed} & Ligand & Pocket \\
\hline
  AlphaFold3  & $\times$   & $\times$   & $\times$   & \checkmark & 1/4 & 9.12 & 1.53 \\
  HighFold3   & $\times$   & $\times$   & $\times$   & \checkmark & 1/4 & 9.82 & 0.98 \\
  Boltz-2     & $\times$   & $\times$   & $\times$   & \checkmark & 1/4 & 1.83 & 0.38 \\
  Protenix-v2 & \checkmark & \checkmark & \checkmark & \checkmark & 4/4 & \textbf{0.75} & 0.41 \\
\hline
\end{tabular}}

\caption{Each model's prediction (plum) for 9EMZ chain E against the deposition (green),
superposed on the receptor and drawn on one shared camera, so the only thing that moves
between the four panels is the predicted peptide. The receptor is the deposition's,
cropped for context. Measurements for each model are tabulated above.}
\label{fig:case-9emz}
\end{figure}

Chain E is a five-residue macrocycle, and four of its five residues sit outside the
standard alphabet: N-methyl-leucine, \emph{allo}-isoleucine, \emph{allo}-threonine, and
a non-standard component \texttt{5BV}, leaving
serine as the only canonical residue. Its bonding is the most ordinary in the benchmark:
all five inter-residue bonds are C--N backbone amides and the ring closes head to tail
through one of them, with no staple, disulfide or thioether involved. Two of the four
models place the peptide (Protenix-v2 0.75\,\AA, Boltz-2 1.83\,\AA) and AF3 and HighFold3 put
it 9.12 and 9.82\,\AA{} away; pocket RMSDs of 1.53 and 0.98\,\AA{} say the binding site
itself was built correctly in both cases, so what went wrong is where the peptide was
placed, which Figure~\ref{fig:case-9emz}a and~c show directly. Only Protenix-v2 passes all
four chemical-validity checks, while the other three each collapse or stretch at least one. 
The residues are unusual here and the bonding is not, and it is the residues the
models struggle with: a five-residue ring held together by nothing but peptide bonds is
arguably the simplest chemistry the benchmark contains, and three of four engines still cannot
build it to a tolerance a linear peptide would meet.

\subsection{8VJP: a correct pose with a broken staple}
\label{sec:case-8vjp}

\begin{figure}[t]
\centering
\includegraphics[width=0.85\linewidth]{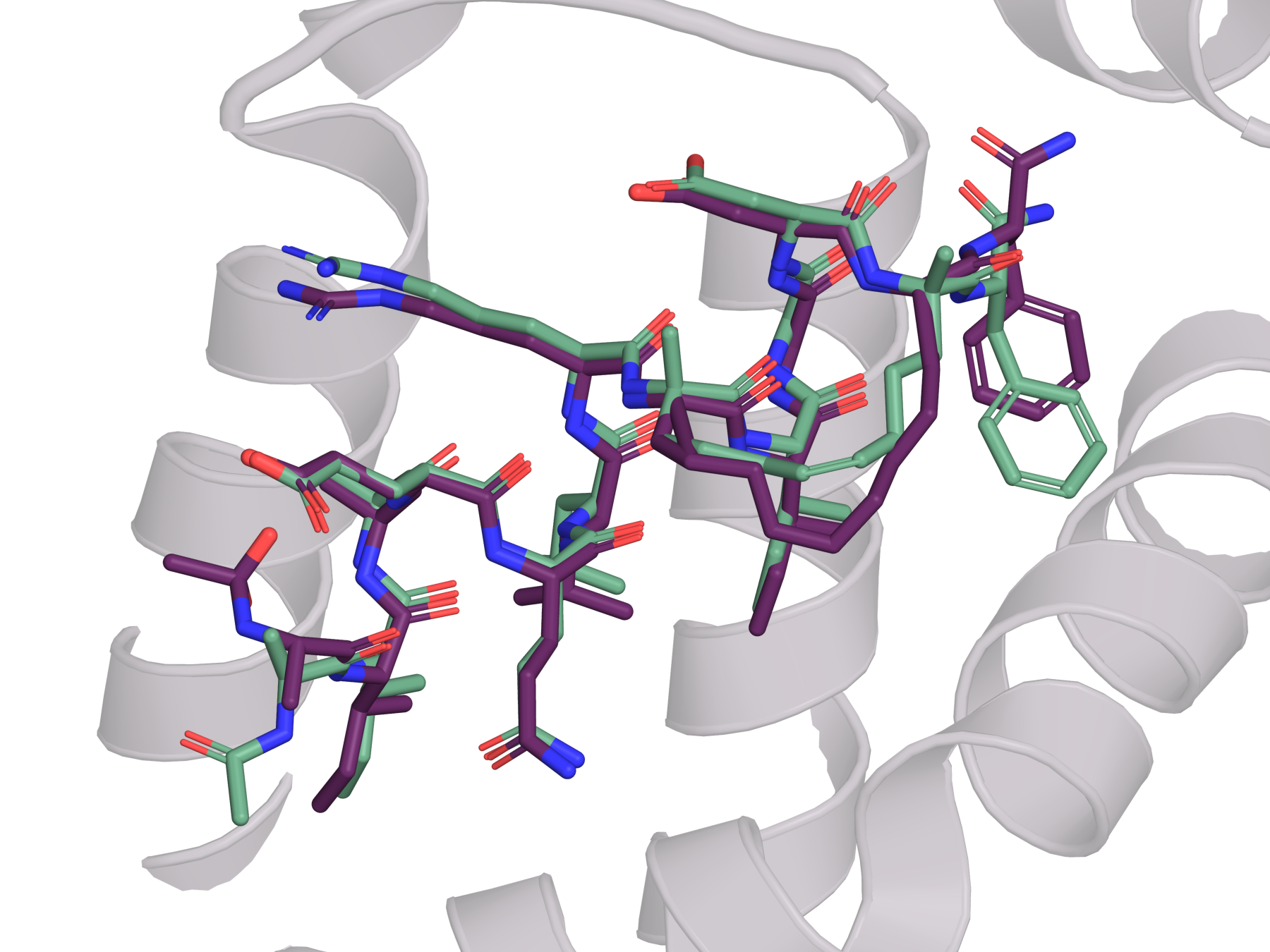}
\caption[Case study: 8VJP chain B]{Protenix (plum) against the deposition (green) for 8VJP
chain B, superposed on the receptor with the deposition's receptor cropped for context.
The peptide is placed correctly and the backbone tracks the deposition; the phenylalanine
ring at the right of the frame is 3.0\,\AA{} out of position.}
\label{fig:case-8vjp}
\end{figure}

Chain B is a fifteen-residue helical peptide stapled between the alpha carbons of \texttt{Ala9} and
\texttt{Ala13} by \texttt{A1AJD}, (4Z)-oct-4-en-1-ol, an all-hydrocarbon crosslink; 
the staple is what makes the molecule a macrocycle, and everything else is a canonical backbone 
with an acetyl cap and a C-terminal amide. All four models dock it, with ligand RMSDs of 1.64 to 
2.00\,\AA, and all four reproduce the helix: the twelve internal C--N backbone amides land within
0.09\,\AA{} of their deposited lengths in every prediction, so on accuracy alone this is an
entry the field would report as solved. The two bonds that attach the crosslink measure
1.58 and 1.55\,\AA{} in the deposition, and every model stretches both --- Boltz-2 to 1.92
and 1.96\,\AA, AF3 to 2.48 and 2.43, HighFold3 to 2.51 and 2.44, Protenix to 2.83 and
2.38. At 2.4\,\AA{} there is no carbon--carbon bond, so three of the four have delivered a
linear peptide alongside an unattached hydrocarbon, and the pose metric cannot reflect so. The
result is a moved side chain: in Figure~\ref{fig:case-8vjp} the core residues
sit within 0.2--0.7\,\AA{} of the deposition while Phe14, the residue immediately after the
second staple anchor, has its phenyl ring 3.0\,\AA{} out (3.7\,\AA{} at the furthest ring
atom) against 1.0--1.5\,\AA{} for its own backbone nitrogen and alpha carbon. The
capping amides are looser than the helix too, Protenix stretching the acetyl cap to
1.55\,\AA{} and Boltz-2 collapsing the C-terminal amide to 0.95\,\AA, and those are what
fail the bond-distance column for those two models. This is the third gap named in
Section~\ref{sec:intro} in a single structure: a prediction inside 2\,\AA{} of the
reference that has distorted the bonds closing the macrocycle past the point where they
are bonds, and reported through pose metrics alone, all four of these predictions are
successes.

\section{Experiments: Correlation between chemical validity and pose accuracy}
\label{sec:val-accuracy}

\begin{figure}[t]
\centering
\includegraphics[width=0.8\textwidth]{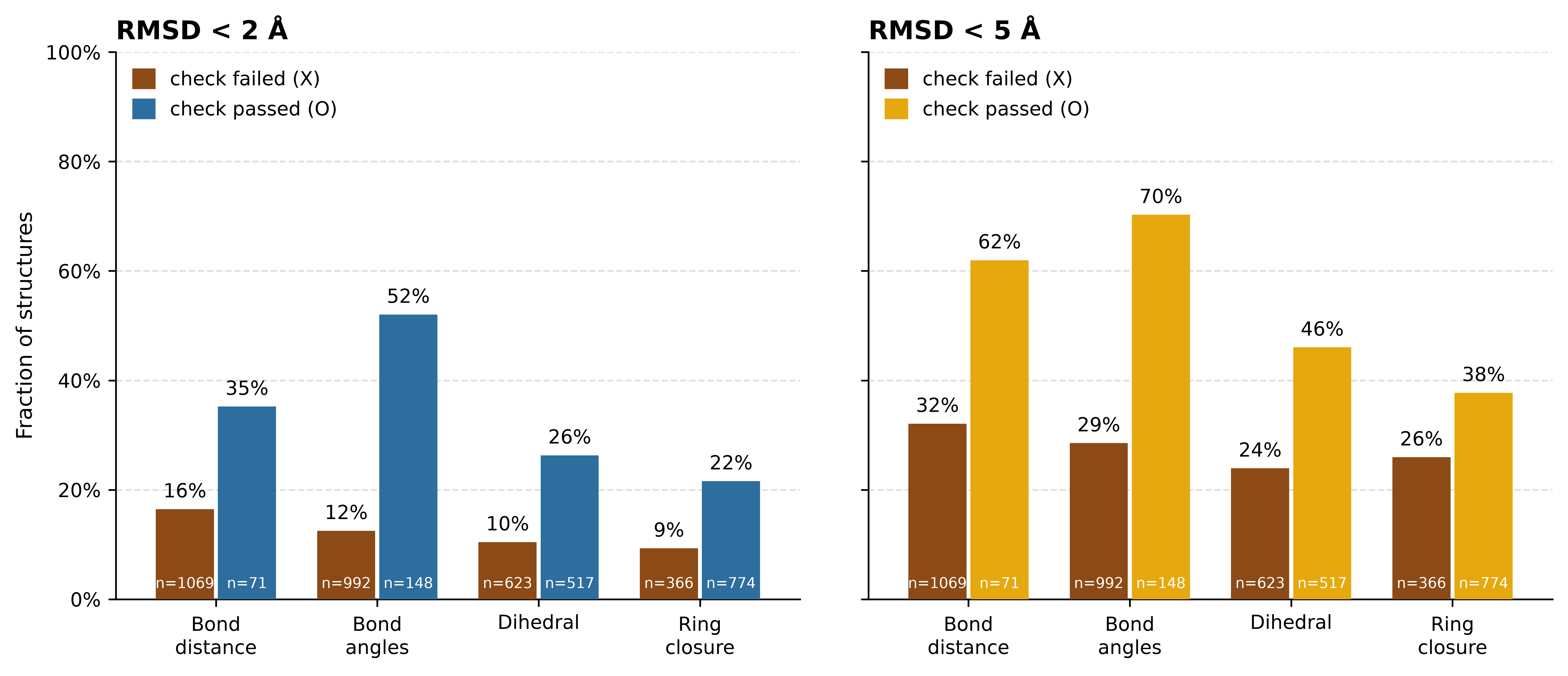}
\caption{Pose accuracy conditioned on chemical validity, pooled over all four engines. For each validity check, bars give the fraction of structures
meeting the $2$\,\AA{} (left) and $5$\,\AA{} (right) RMSD thresholds among predictions that fail (brown) and pass (blue/gold) that check; the dashed line is the unconditional success rate and $n$ is the number of structures in each group.}
\label{fig:validity-vs-rmsd}
\end{figure}

We took each of the 1,140 predictions (285 peptides × 4 models), recorded whether it passed each of the four chemical validity checks, and asked how well those four pass/fail flags predict whether the prediction is also geometrically accurate — within 2 \AA or 5 \AA of the deposited structure (Figure \ref{fig:validity-vs-rmsd}). We can see there is a clear correlation between chemical validity and pose accuracy, showing up to 51\% of difference between success rate. However we wanted to use the logistic model enters all four at once, so each coefficient is the effect of that check among predictions that are otherwise alike. The result is that only two of the four carry real information. Bond angles dominate: passing that check multiplies the odds of a sub-2 \AA pose by about five. Dihedrals contribute a smaller but solid factor of roughly two. Bond distance turns out to carry nothing on its own — its apparent effect in the figure is borrowed entirely from its correlation with angles — and ring closure is weak and borderline. But the striking part is how little the four checks explain overall: pseudo-R² of about 0.12, and a structure that fails every single check still has a roughly 8\% chance of landing within 2 \AA and a quarter chance of landing within 5 \AA. So chemical validity and geometric accuracy are measuring largely different things. A model can build a chemically sane molecule in the wrong place, or the right shape out of broken chemistry, and neither metric alone tells you whether you have a usable macrocycle prediction — which is the argument for reporting both.

\section{Experiments: Correlation between confidence and pose accuracy/chemical validity}
\label{sec:confidence}

\paragraph{Confidence against pose accuracy.}
\begin{figure}[t]
\centering
\includegraphics[width=1\textwidth]{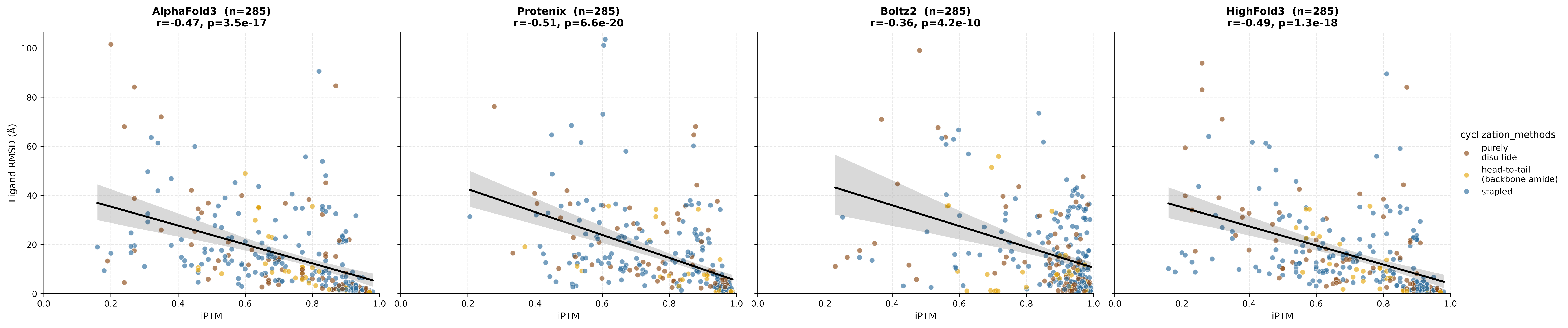}
\caption{Model confidence versus pose accuracy on the 285 targets predicted by all four engines. Each point is one target, colored by cyclization chemistry; the line is
an ordinary least-squares fit with a 95\% confidence band, and Pearson $r$ with its
two-sided $p$-value is given per panel. }
\label{fig:iptm-rmsd}
\end{figure}

Figure~\ref{fig:iptm-rmsd} relates each engine's self-reported interface predicted TM-score (ipTM) of the entire predicted complex to the pocket-aligned ligand RMSD achieved on the 285 targets. Across all four models, confidence is moderately predictive of accuracy across the board: Pearson correlations range from $r=-0.36$ (Boltz-2) to $r=-0.51$ (Protenix-v2), all highly significant ($p<10^{-10}$), with the ordering broadly matching the accuracy ranking of Table~\ref{tab:accuracy-validity}. At these
magnitudes, ipTM accounts for roughly a quarter of the variance in RMSD, so it is useful
for coarse triage but not a reliable per-target indicator of success; a substantial
number of predictions above ipTM $0.8$ still sit well beyond $5$\,\AA. Importantly, the ipTM reported here is a global metric. Therefore, the confidence values can be boosted by complexes with multiple receptor chains, even in cases where the peptide is incorrectly modeled.

\paragraph{Confidence against chemical validity.}
The same question can be asked of chemical validity. Each chain--model pair carries one
pass/fail flag per category, and Table~\ref{tab:iptm-r} gives the correlation between ipTM and
each flag. Against pose accuracy ipTM correlates at $r = -0.36$ to $-0.51$; against chemical
validity it is far weaker, and weakest of all on ring closure, where no engine reaches
significance ($r = +0.035$ to $+0.094$, $p = 0.12$--$0.55$). Bond angles are the only category significant for all four engines ($p = 1.4 \times 10^{-7}$--$0.037$). Confidence therefore carries signal about local geometry
and almost none about whether the macrocycle was closed.

\begin{table}[h]
\centering
\caption{Pearson correlation between ipTM and each chemical validity check ($n = 285$ per model).}
\label{tab:iptm-r}
\begin{tabular}{|l|c|c|c|c|}
\hline
Model & Bond distance & Bond angles & Dihedral & Ring closure \\
\hline
AlphaFold3  & 0.126 & 0.199 & 0.070 & 0.060 \\
HighFold3   & 0.094 & 0.144 & 0.062 & 0.036 \\
Boltz-2     & 0.067 & 0.024 & -0.014 & 0.018 \\
Protenix-v2 & 0.164 & 0.306 & 0.209 & 0.094 \\
\hline
\end{tabular}
\end{table}

\section{Experiments: Five Sample Experiments}
\label{sec:additional-experiment}

Every method emits five samples per target, but a user receives one: the sample
the method's own confidence score ranks first. This section separates what the
methods can produce from what they deliver.

\paragraph{Protocol.}
All five samples of every target were scored with the pocket-aligned peptide
RMSD of the main text, using the identical correspondence and alignment
procedure. One target (9DZJ\_C) has a single Boltz-2 output and was dropped from
all methods, leaving 284 targets $\times$ 4 methods $\times$ 5 samples $=5680$
evaluations; none failed. Three summaries are reported: \emph{pooled}, treating
all five samples as independent observations; \emph{top-ranked}, the sample the
method's confidence score selects, which is what a user actually obtains; and
\emph{best-of-5}, the lowest RMSD among the five, an oracle that bounds what
perfect sample selection could achieve.

\begin{table}[htbp]
  \centering
  \caption{Pooled over all five samples ($n=1420$ per method).}
  \label{tab:pooled}
  \begin{tabular}{lrrrr}
    \hline
    Method & $n$ & Median & $<$2~\AA & $<$5~\AA \\
    \hline
    AlphaFold3  & 1420 & 10.03         & 242 (17.0\%)          & 458 (32.3\%) \\
    Boltz-2     & 1420 & 10.06         & 210 (14.8\%)          & 478 (33.7\%) \\
    Protenix-v2 & 1420 & \textbf{8.94} & \textbf{285} (20.1\%) & \textbf{548} (38.6\%) \\
    HighFold3   & 1420 & 10.34         & 260 (18.3\%)          & 465 (32.7\%) \\
    \hline
    All         & 5680 & 9.83          & 997 (17.6\%)          & 1949 (34.3\%) \\
    \hline
  \end{tabular}
\end{table}

\begin{table}[htbp]
  \centering
  \caption{Delivered versus attainable accuracy ($n=284$ per method). Best-of-5
    is an oracle and is not achievable in practice.}
  \label{tab:selection}
  \begin{tabular}{lrrr rrr}
    \hline
    & \multicolumn{3}{c}{Top-ranked (delivered)} & \multicolumn{3}{c}{Best-of-5 (oracle)} \\
    \cline{2-4}\cline{5-7}
    Method & Median & $<$2~\AA & $<$5~\AA & Median & $<$2~\AA & $<$5~\AA \\
    \hline
    AlphaFold3  & 10.19 & 50 (17.6\%)          & 89 (31.3\%)           & 8.73          & 55 (19.4\%)          & 101 (35.6\%) \\
    Boltz-2     & 10.07 & 42 (14.8\%)          & 98 (34.5\%)           & 8.36 & 48 (16.9\%)          & 105 (37.0\%) \\
    Protenix-v2 & \textbf{8.91} & \textbf{55} (19.4\%) & \textbf{107} (37.7\%) & \textbf{7.59} & \textbf{66} (23.2\%) & \textbf{121} (42.6\%) \\
    HighFold3   & 9.81  & 54 (19.0\%)          & 93 (32.7\%)           & 8.64          & 61 (21.5\%)          & 105 (37.0\%) \\
    \hline
  \end{tabular}
\end{table}

\paragraph{Sampling is tight for most targets and unstable for a quarter of them.}
Across the five samples of a given target the median within-target standard
deviation is 0.53--0.66~\AA{} and the median spread between best and worst
sample is 1.24--1.64~\AA, so for most targets the choice of sample is
immaterial. But 20--24\% of targets have a within-target standard deviation
above 2~\AA: the five samples disagree about where the peptide sits. A
single-sample number conceals this bimodality entirely.

\paragraph{Confidence ranking is indistinguishable from random selection.}
Because the top-ranked prediction is always one of the five samples, the relevant null is that the delivered sample is drawn uniformly from them. Under that null the expected number of sub-2~\AA{} predictions is
$\sum_i k_i / 5$, where $k_i$ is the number of a target's five samples meeting
the threshold. Pooled across methods the expectation is $199.4$ against an
observed $198$ ($z=-0.38$, $p=0.71$); no method departs from it individually
(AlphaFold3 $50$ vs $48.4$, $p=0.74$; HighFold3 $55$ vs $52.0$, $p=1.00$;
Boltz-2 $42$ vs $42.0$, $p=1.00$; Protenix-v2 $55$ vs $57.0$, $p=0.29$).
Confidence-based selection adds nothing on this benchmark, consistent with the
weak association between ipTM and chemical validity reported earlier.

\paragraph{Perfect selection would not close the gap.}
Best-of-5 raises the sub-2~\AA{} count from 201 to 230 across all methods, from
$17.4\%$ to $20.2\%$, and the sub-5~\AA{} count from 387 to 432. An oracle with
advance knowledge of the answer therefore recovers fewer than three percentage
points. The failures are not cases where a correct pose was generated and then
mis-ranked; in the large majority of targets no correct pose is produced in five
attempts. Method ordering is unchanged under every selection rule, with
Protenix-v2 leading and Boltz-2 trailing at the 2~\AA{} threshold.

\section{Bond Distance Analysis by Bond Type}
\label{sec:bond-forest-experiment}

\begin{figure}[t]
\centering
\includegraphics[width=0.8\textwidth]{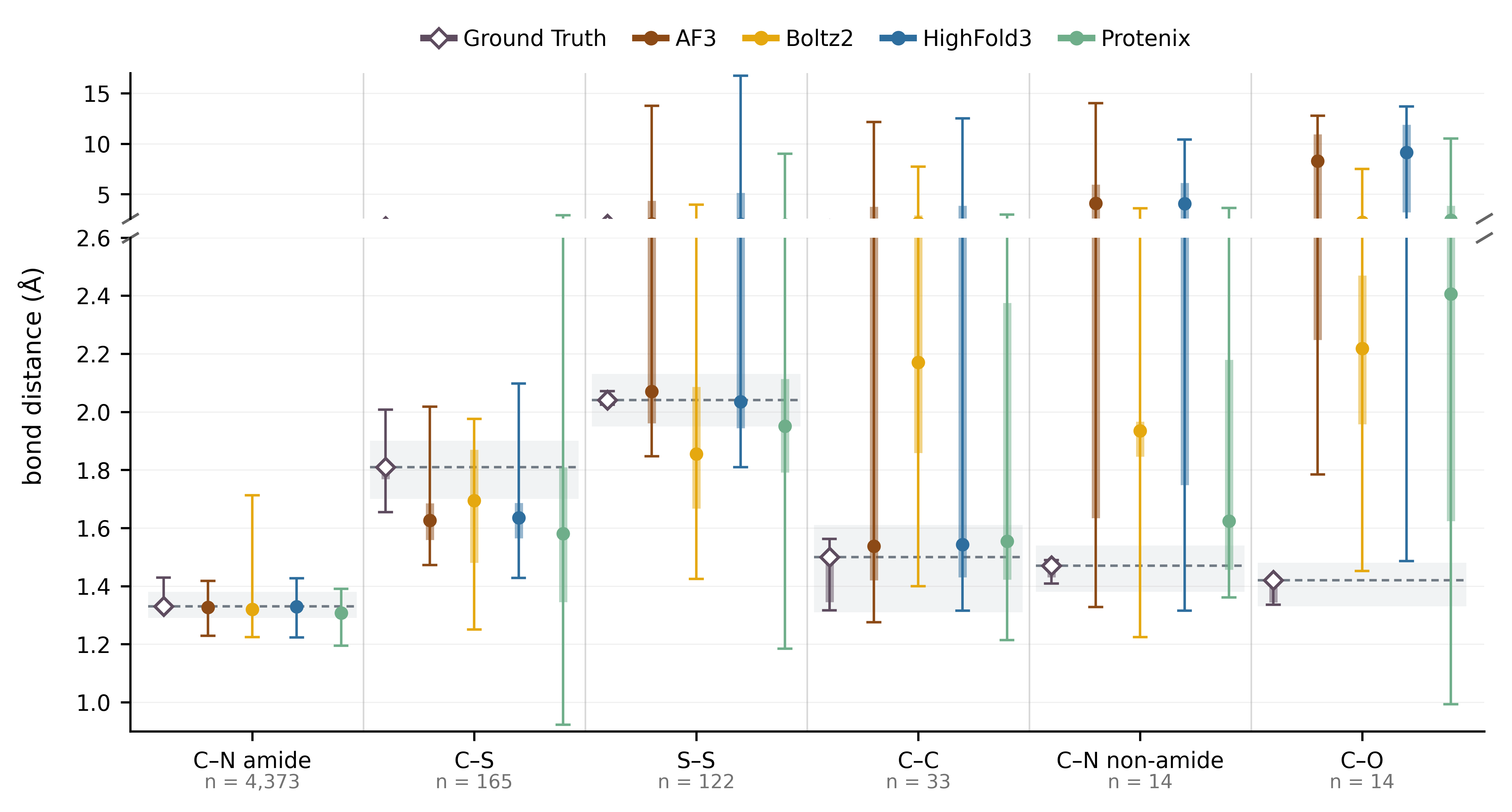}
\caption{Bond distances by bond type for the ground truth and the four co-folding models. Markers show the median, thick bars the 25th--75th percentile range. Dashed lines and shaded bands mark the ground-truth median and reference interval. $n$ is the number of bonds of each type. The y-axis is broken at 2.6\,\AA.}
\label{fig:forest_bond}
\end{figure}

We additionally analyzed bond distances by bond type (Figure~\ref{fig:forest_bond}). For C--N amide bonds, which make up the large majority of bonds ($n = 4{,}373$), all four models closely match the ground truth, with narrow distributions. The other bond types deviate in two ways. First, some are systematically shifted: predicted C--S bonds are shorter than the ground truth for every model (medians of 1.58--1.70\,\AA{} against 1.81\,\AA). Second, the distributions have heavy tails. For S--S and C--C bonds, the medians of most models lie near the ground truth, but the upper whiskers extend to several \aa ngstr\"oms, which corresponds to bonds that are not formed at all. The deviation is largest for C--N non-amide and C--O bonds, where the AlphaFold3 and HighFold3 medians themselves lie at roughly 4\,\AA{} and 8--9\,\AA, respectively, although these two types are rare ($n = 14$ each). Overall, the models reproduce the backbone amide geometry well but are unreliable for the bond types that define cyclization and non-canonical chemistry.

\section{Covariate-Adjusted Statistical Analysis}
\label{sec:statistical-analysis}

The comparisons across cyclization classes and configuration families in Section~6.2 are descriptive. Here we test whether they survive adjustment for covariates that differ between groups, and whether the engines differ in those effects.

\paragraph{Design and model.}
PRETZEL-SET is a complete balanced design: all 285 peptide chains are predicted by all four engines, giving 1{,}140 observations with no missing values. For peptide $i$ under engine $j$ we fit
\begin{equation}
\begin{aligned}
\log \mathrm{RMSD}_{ij} ={}& \beta_0 + \alpha_j + \gamma_{\mathrm{cyc}(i)}
 + \beta_L\,\mathrm{length}_i + \beta_N\,\mathrm{ncAA}_i \\
&+ \beta_S\,\mathrm{segments}_i + \beta_E\,\mathrm{entities}_i
 + \beta_B\,\mathrm{bonds}_i + \beta_M\,\mathrm{multichain}_i
 + u_i + \varepsilon_{ij},
\end{aligned}
\end{equation}
with $u_i \sim \mathcal{N}(0,\tau^2)$ and $\varepsilon_{ij} \sim \mathcal{N}(0,\sigma^2)$. Categorical terms are treatment-coded (AlphaFold3 and head-to-tail as references) and numeric covariates are mean-centred, so $\beta_0$ is the expected log RMSD for an AlphaFold3-predicted head-to-tail macrocycle of average size and complexity (1.77, or 5.9\,\AA). The response is logged because the raw-scale residuals are strongly skewed (skewness $1.86 \rightarrow -0.18$; Shapiro--Wilk $W = 0.855 \rightarrow 0.977$). Coefficients are therefore multiplicative, with $\exp(\beta)-1$ the percentage change in RMSD.

Because every covariate is constant within a peptide, engine interactions are identified entirely within peptide. We therefore test the seven interaction blocks reported below against the engine-only model with peptide fixed effects absorbed, using an exact within-peptide $F$ test with Holm correction and confirming each result with a 10{,}000-sample Freedman--Lane permutation test. Pooled OLS is invalid for these tests: it returns $p = 0.99$ on the omnibus test where the exact within-peptide test returns $p = 0.002$.

\begin{table}[t]
\centering
\caption{Mixed-effects model of log RMSD (ML fit, $n = 1{,}140$, 285 peptide groups; $\tau^2 = 1.02$, $\sigma^2 = 0.28$).}
\label{tab:mixed-model}
\vspace{0.5em}
\small
\setlength{\tabcolsep}{5pt}
\begin{tabular}{@{}l cccc@{}}
\hline
Term & $\beta$ & SE & $\exp(\beta)-1$ & $p$ \\
\hline
Intercept & 1.771 & 0.222 & -- & $<0.001$ \\
Engine: Boltz-2 & $+0.036$ & 0.044 & $+3.6\%$ & 0.42 \\
Engine: HighFold3 & $-0.034$ & 0.044 & $-3.3\%$ & 0.45 \\
Engine: Protenix-v2 & $-0.131$ & 0.044 & $-12.2\%$ & 0.003 \\
Cyclization: purely disulfide & $+0.238$ & 0.256 & $+26.9\%$ & 0.35 \\
Cyclization: stapled & $+0.350$ & 0.216 & $+42.0\%$ & 0.11 \\
Length (per residue) & $+0.016$ & 0.010 & $+1.6\%$ & 0.10 \\
ncAA count (per residue) & $+0.025$ & 0.023 & $+2.6\%$ & 0.27 \\
Chain segments & $-0.365$ & 0.348 & $-30.6\%$ & 0.29 \\
Bonded ligand entities & $-0.240$ & 0.133 & $-21.4\%$ & 0.07 \\
Covalent cross-links & $+0.376$ & 0.112 & $+45.6\%$ & 0.0008 \\
Multi-chain & $+0.204$ & 0.473 & $+22.6\%$ & 0.67 \\
\hline
\end{tabular}
\end{table}

Table~\ref{tab:mixed-model} reports the coefficients. Cross-link count is the only chemistry term that reaches significance. The negative coefficient on bonded ligand entities is conditional and does not mean that entities help: entity count correlates with cross-link count at $r = 0.84$, and fitted alone its effect is positive ($+20.5\%$, $p = 0.003$), matching Figure~4. The two cannot be separated at this sample size.

Adjustment is necessary because the covariates are confounded with cyclization class. Head-to-tail peptides are short and ncAA-rich (10.3 residues, 4.4 ncAAs), disulfide peptides are long and nearly canonical (23.6 residues, 0.9 ncAAs), and stapled peptides carry the most cross-links (3.0 against 1.7).

\paragraph{Difficulty is shared across engines.}
Peptide identity accounts for 80.5\% of the variance in log RMSD (intraclass correlation), while engine identity explains 0.26\% ($R^2$ of an engine-only model). The engines agree on which peptides are hard (pairwise Spearman $\rho = 0.78$--$0.93$), and 74.7\% of peptides receive the same verdict from all four engines: 65 are solved by all at $<5$\,\AA, 148 are failed by all, and 72 are contested. The engine $\times$ configuration-family interaction is not significant ($F(12, 840) = 1.14$, $p = 0.32$; permutation $p = 0.33$), and the between-engine Spearman correlation over the five family means is 0.9--1.0. The engine $\times$ cyclization-class interaction is the closest to significance of the seven blocks ($p = 0.017$ uncorrected, $0.10$ after Holm), and the ranking does reverse: AlphaFold3 is the weakest engine on head-to-tail targets and the strongest on disulfide-only targets.

\paragraph{No evidence of super-additivity.}
The single-chain families form a $2 \times 2$ design (inline CCD modification $\times$ bonded ligand entity) over the 255 peptides that remain after setting aside the 30 multi-chain cases. The CCD-plus-ligand cell is the worst of the four (9.6\% under 2\,\AA{} against 26.5\% for plain single chains), but the interaction is not significant in any specification (Table~\ref{tab:interaction}). The unadjusted estimate points towards super-additivity, but it is well within noise and shrinks towards zero as covariates are added.

\begin{table}[t]
\centering
\caption{Interaction between inline CCD modification and bonded ligand entity ($n = 255$ single-chain peptides). $\beta > 0$ and $\mathrm{OR} < 1$ both indicate super-additivity.}
\label{tab:interaction}
\vspace{0.5em}
\small
\setlength{\tabcolsep}{5pt}
\begin{tabular}{@{}l cc@{}}
\hline
Outcome and model & Interaction estimate & $p$ \\
\hline
log RMSD, unadjusted & $\beta = +0.246$ & 0.38 \\
log RMSD, + length, ncAA & $\beta = +0.062$ & 0.83 \\
log RMSD, + cyclization class, cross-link count & $\beta = +0.112$ & 0.71 \\
$P(\mathrm{RMSD} < 2\,\text{\AA})$, unadjusted & $\mathrm{OR} = 0.59$ & 0.36 \\
$P(\mathrm{RMSD} < 2\,\text{\AA})$, fully adjusted & $\mathrm{OR} = 1.12$ & 0.88 \\
\hline
\end{tabular}
\end{table}

\paragraph{Cross-link count as the main driver.}
The strongest chemistry predictor is the number of declared covalent cross-links, at $+46\%$ RMSD per additional cross-link ($p = 0.0008$). Once it is included, the cyclization-class effect weakens substantially (Table~\ref{tab:stapled}). Cross-link count and the stapled label are correlated ($r = 0.47$) but separable ($\mathrm{VIF} \leq 5.95$ in the full model). Cyclization class alone explains under 3\% of the variance in log RMSD ($R^2 = 0.027$), and the ncAA-count effect does not survive adjustment ($p = 0.27$).

\begin{table}[t]
\centering
\caption{Effect of stapled versus head-to-tail cyclization on RMSD under increasing adjustment.}
\label{tab:stapled}
\vspace{0.5em}
\small
\setlength{\tabcolsep}{5pt}
\begin{tabular}{@{}l cc@{}}
\hline
Model & Stapled vs.\ head-to-tail & $p$ \\
\hline
Cyclization alone & $+88\%$ & 0.002 \\
+ length, ncAA & $+75\%$ & 0.007 \\
+ structural complexity & $+42\%$ & 0.106 \\
\hline
\end{tabular}
\end{table}

\paragraph{Limitation.}
Architecture, training data, and input specification are bundled: each engine is one fixed combination of the three. Separating them would require retraining an architecture on controlled corpora or running an engine under varied input specifications, neither of which is possible from published checkpoints. Our results therefore bound the total engine effect without apportioning it.

\end{document}